\documentclass{article}

\usepackage[preprint]{neurips_2025}

\usepackage[utf8]{inputenc} 
\usepackage[T1]{fontenc}    
\usepackage{hyperref}       
\usepackage{url}            
\usepackage{booktabs}       
\usepackage{amsfonts}       
\usepackage{nicefrac}       
\usepackage{microtype}      
\usepackage{xcolor}         
\usepackage{graphicx}
\usepackage[table]{xcolor}
\usepackage{amsmath}
\usepackage{amsthm}
\usepackage{amssymb}
\usepackage{multirow}
\hypersetup{
    colorlinks=true,
    linkcolor=red,
    citecolor=teal,
    urlcolor=cyan,
}
\newtheorem{proposition}{Proposition}
\newtheorem{definition}{Definition}
\usepackage{algorithmic}
\usepackage{algorithm}
\usepackage{wrapfig}

\title{Per-parameter Task Arithmetic for Unlearning in Large Language Models}

\author{%
  \textbf{Chengyi Cai}$^1$, \textbf{Zesheng Ye$^1$}, \textbf{Jiangchao Yao$^2$}, \textbf{Jianzhong Qi$^1$}, \\
  \textbf{Bo Han$^3$}, \textbf{Xiaolu Zhang$^4$}, \textbf{Feng Liu$^1$}, \textbf{Jun Zhou$^4$} \\
  \quad$^1$The University of Melbourne
  \quad$^2$Shanghai Jiao Tong University \\
  \quad$^3$Hong Kong Baptist University
  \quad$^4$Ant Group \\
  \texttt{fengliu.ml@gmail.com}
}

\begin{document}

\maketitle

\begin{abstract}
In \textit{large language model} (LLM) unlearning, private information is required to be removed. Task arithmetic unlearns by subtracting a specific \textit{task vector} (TV)--defined as the parameter difference between a privacy-information-tuned model and the original model. While efficient, it can cause over-forgetting by disrupting parameters essential for retaining other information. 
Motivated by the observation that each parameter exhibits different importance for forgetting versus retention, we propose a \textit{per-parameter task arithmetic} (PerTA) mechanism to rescale the TV, allowing per-parameter adjustment. These weights quantify the relative importance of each parameter for forgetting versus retention, estimated via \textit{\textbf{grad}ients} (i.e., PerTA-grad) or the \textit{diagonal \textbf{Fisher} information approximation} (i.e., PerTA-fisher). Moreover, we discuss the effectiveness of PerTA, extend it to a more general form, and provide further analysis.
Extensive experiments demonstrate that PerTA consistently improves upon standard TV, and in many cases surpasses widely used training-based unlearning methods in both forgetting effectiveness and overall model utility. By retaining the efficiency of task arithmetic while mitigating over-forgetting, PerTA offers a principled and practical framework for LLM unlearning.

\end{abstract}

\section{Introduction}
\begin{figure}[htbp]
\centering
\begin{minipage}{0.4\textwidth}
  \centering
  \includegraphics[width=\linewidth]{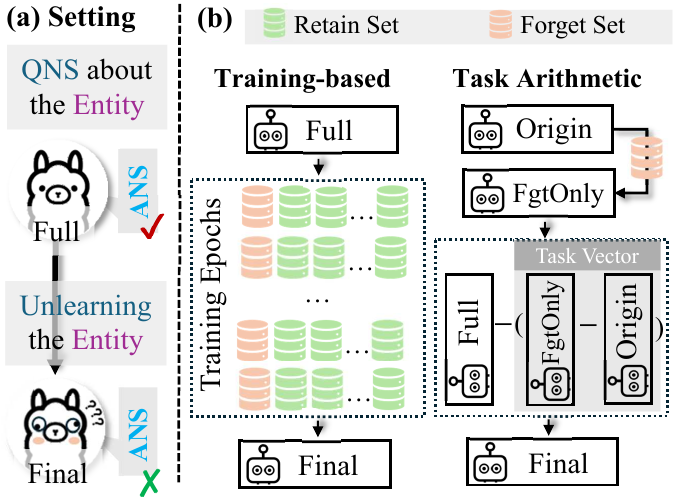}
  \vspace{-8pt}
    \caption{The task of LLM unlearning and mainstream method categories. (a) depicts the problem setting, where the objective is to erase knowledge of specific entities. (b) contrasts training-based approaches with task arithmetic.}
    \label{fig:intro}
\end{minipage}
\hfill
\begin{minipage}{0.57\textwidth}
  \centering
  \includegraphics[width=\linewidth]{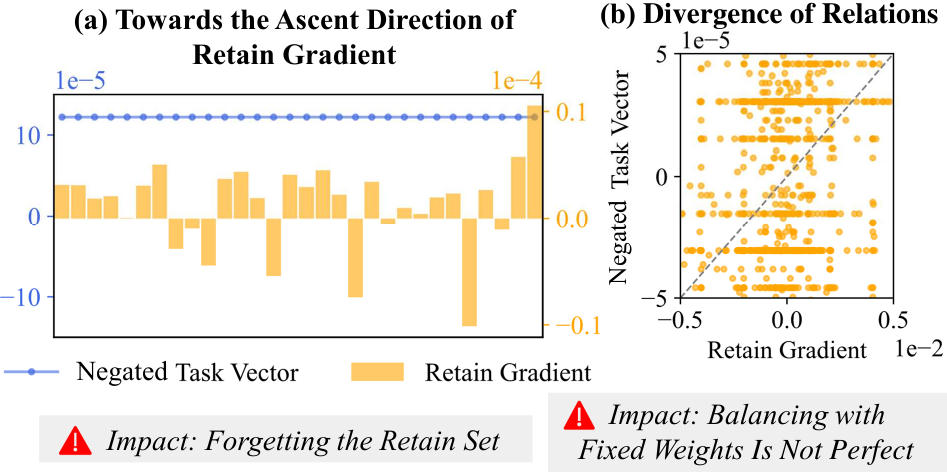}
  \vspace{-8pt}
    \caption{Bottlenecks of task arithmetic methods. (a) illustrates that TV may steer the model toward the ascent direction of the retained gradient, leading to over-forgetting. (b) shows per-parameter divergence of TV--retain gradient relations, rendering the problem non-trivial and not solvable by a uniform weight.}
    \label{fig:motiv}
\end{minipage}
\vspace{-10pt}
\end{figure}

\textit{Large language models} (LLMs) can continually acquire new knowledge~\citep{lu2024spp,luo2024wizardarena} through post-training; however, the integration of newly ingested data may raise concerns regarding privacy, intellectual property, or misinformation~\citep{karamolegkou2023copyright}. Due to their tendency to memorize training data, LLMs may inadvertently disclose sensitive information when queried. LLM unlearning~\citep{liu2025rethinking,yao2024large} aims to erase the memory of specified entities from LLMs to mitigate such risks, as shown in Figure~\ref{fig:intro}(a).

Some \textit{training-based} LLM unlearning methods achieve forgetting of specific entities (i.e., forget set) by designing carefully crafted unlearning loss functions~\citep{NPO,SimNPO,yao2024large,satimp} and incorporating entities to be retained (i.e., retain set) to ensure that unrelated knowledge in the model remains unaffected~\citep{GD}. Another counterpart, \textit{task arithmetic}, avoids multiple iterative training epochs with extensive data, as illustrated in Figure~\ref{fig:intro}(b), where full model and final model represent LLMs before and after unlearning, respectively. This approach achieves unlearning by subtracting from the full model a specific \textit{task vector} (TV)~\citep{TV} for the forget set. TV denotes the parameter difference between a model finetuned solely on the forget set (hereafter referred to as the FgtOnly model) and the original pretrained model (hereafter the Origin model).

However, the potential correlation and coupling between the entity to be unlearned and other knowledge may cause the subtracted task vector to also contain changes in parameters crucial for preserving other knowledge, thereby risking excessive forgetting of entities that should be retained. Figure~\ref{fig:motiv}(a) takes the task of unlearning 1\% of entities from the TOFU~\citep{tofu} dataset as an example. The top 30 parameters with the largest values in the negated task vector (i.e., $-V$, which is added to the full model $\theta_{\rm full}$ to obtain the final model $\theta_{\rm final}=\theta_{\rm full}+(-V)$) were selected as examples. For these parameters, we plotted both the values in the negated TV and the gradient magnitudes with respect to the retain set (i.e., the retain gradient) of the same parameters. The results show that, for most of these parameters, the direction indicated by the negated TV aligns with the gradient ascent direction for the retain set. This implies that directly adding the negated TV to the full model would lead to forgetting of the entities that are supposed to be retained. A simple remedy is to add a uniform weight $\omega \in \mathbb{R}$ satisfying $0<\omega<1$ to TV to reduce the effect of TV (i.e., $\theta_{\rm final}=\theta_{\rm full}+\omega\cdot(-V)$), thereby balancing between unlearning and retaining. However, as shown in Figure~\ref{fig:motiv}(b), we find that such an approach may be suboptimal as it ignores per-parameter divergence. By plotting the negated TV and the retain gradient corresponding to different parameters, we observe that different parameters exhibit varying relations between TV and retain gradients, requiring a more sophisticated paradigm.

After formulating the problem, in Section~\ref{sec:pre}, we propose the \textit{\textbf{Per}-parameter \textbf{T}ask \textbf{A}rithmetic} (PerTA) mechanism as a solution to the aforementioned bottleneck. PerTA assigns different weights to each parameter in TV and performs a per-parameter multiplication (i.e., $\theta_{\rm final}=\theta_{\rm full}+W\odot (-V)$) to flexibly control the magnitude of editing, where $W$ is a matrix with the same size as $\theta_{\rm full}$.
Parameters that are more pivotal for unlearning can be assigned higher weights, whereas those crucial for retention receive lower weights, aiming to facilitate both unlearning and retention.

In Section~\ref{sec:method}, we detail how per-parameter weights are estimated using absolute \textit{\textbf{grad}ients} (which captures the importance of parameters given forget or retain sets, abbreviated as PerTA-grad) or the diagonal \textit{\textbf{Fisher} information approximation} (which reflects the sensitivity of parameters to forget or retain sets, abbreviated as PerTA-fisher). We analyze its effectiveness by defining \textit{retain-forget ratio} for parameters.
We also extend PerTA to a generalized form and provide further discussion.

In Section~\ref{sec:exp}, we evaluate PerTA on two commonly used unlearning benchmarks TOFU~\citep{tofu} and MUSE~\citep{muse} across multiple metrics. Results show that PerTA not only substantially outperforms its baseline, vanilla TV, but also exceeds the performance of several mainstream training-based unlearning methods. Training time analysis confirms that PerTA is efficient, while qualitative results illustrate its ability to retain knowledge upon effective unlearning.

PerTA extends TV by preserving retention while enabling effective unlearning, all with high efficiency. Remarkably, it achieves performance surpassing several training-based unlearning methods, highlighting its practical effectiveness. Beyond empirical gains, PerTA offers a new task-arithmetic perspective for LLM unlearning research and introduces a flexible approach for balancing modification and retention in LLM task arithmetic.

\section{Related Works}
\textbf{LLM Unlearning.}
Machine unlearning~\citep{unlearning1,unlearning2,unlearning3,unlearning4,lu2022quark} aims to selectively remove some previously acquired knowledge from a model while preserving its overall utility. 
LLM unlearning has attracted increasing attention, playing a vital role in correcting misinformation, mitigating biases, and protecting privacy~\citep{fantowards,yao2024survey,jang2022knowledge}. Recent studies on LLM unlearning have advanced this field from multiple perspectives, including benchmarks~\citep{tofu,muse,wmdp}, frameworks~\citep{openunlearning}, evaluation protocols~\citep{wangrethinking,wangtowards}, methodological innovations~\citep{jia2024soul,pawelczykcontext,kadhe2024split}, {and hallucination mitigation~\citep{shen2025lunar,zhang2025rule}. Different objectives and problem settings of unlearning are discussed in Appendix~\ref{app:obj}}

Among training-based unlearning methods, GA~\citep{yao2024large} is the pioneering work that minimizes the log-likelihood of the entities to be unlearned. GD~\citep{GD} improves it by incorporating the loss on a retain set to mitigate forgetting. NPO~\citep{NPO} constructs its loss function by separating the dis-preferred component from DPO~\citep{dpo}, while SimNPO~\citep{SimNPO} further removes the reliance on reference models. GRU~\citep{GRU} projects the unlearning gradient onto the orthogonal space of retain gradients, and SatImp~\citep{satimp} reweights the loss on a token-wise basis.
\citealp{muse} introduces TV~\citep{TV} into the unlearning setting. Despite the rapid progress of training-based methods, challenges remain in terms of time and data efficiency, motivating the exploration of more efficient alternatives such as task arithmetic. Since these methods are currently underexplored, we aim to investigate the potential of task arithmetic-based methods.

\textbf{Model Merging.}
Model merging, also referred to as model editing, is a cost-effective approach that directly manipulates the weight space of multiple pretrained models. \citealp{TV} introduces the concept of TV, defined as the difference between a finetuned model on a given task and its original counterpart, which can then be used for subsequent model merging. \citealp{tangent} further investigates the fundamental mechanisms of TV by analyzing linearized models. AdaMerging~\citep{adamerging} improves upon the TV framework by learning task-wise or layer-wise coefficients, enabling more effective multi-task learning. {Additional refinements include trimming~\citep{dare}, sign selection~\citep{ties} before merging, and composing parameter blocks~\citep{atlas} or models~\citep{lee2025dynamic} with learned coefficients.}

Recently, {model merging} has been successfully extended to LLMs~\citep{metagpt,fusellm,fusionchat} and multimodal LLMs~\citep{mllm1,mllm2}. Within the context of LLMs, MetaGPT~\citep{metagpt} employs a task arithmetic approach that exploits the local linearity of LLMs together with the approximate orthogonality of TVs. FuseLLM~\citep{fusellm} and FusionChat~\citep{fusionchat} investigate strategies for integrating multiple pretrained LLMs in the parameter space to obtain a more potent model. While existing studies have primarily focused on multi-task learning scenarios, our paper explores model merging in LLM unlearning, along with potential improvements. {Unlike other model merging methods that combine knowledge, we study task arithmetic in this paper to remove knowledge from the pretrained models.}
\section{Preliminaries and Insights}
\label{sec:pre}
We consider a pretrained auto-regressive LLM parameterized by $\theta_0$ with self-attention structures~\citep{liu2018generating}.
In the post-training phase, the LLM can be finetuned on new knowledge $\mathcal D=\{s^1,s^2, ..., s^{|\mathcal D|}\}$ consisting of $|\mathcal D|$ sequences, where each sequence $s=[t_1, t_2, ..., t_{|s|}]$ contains $|s|$ tokens. 
Denoting $t_{<i}$ as the subsequence of $s$ from $t_1$ to $t_{i-1}$, the probability of $s$ given parameter $\theta$ can be defined as $p(s;\theta)\triangleq \prod_{i=1}^{|s|}p(t_i|t_{<i};\theta)$, which is the product of the conditional probabilities of all tokens. Then $\theta$ can be learned by minimizing the negative log likelihood loss: 
\begin{equation}\label{eq:loss}
   \mathcal L(\mathcal D;\theta)=-\frac{1}{|\mathcal D|}\sum\nolimits_{s \in \mathcal D}\log p(s;\theta).
\end{equation}
Given a new target knowledge set $\mathcal D_{\rm full}$, the finetuned model $\theta_{\rm full}$ on the whole dataset (i.e., the full model) can be obtained by the training objective $\arg\min_{\theta \in \Theta}\mathcal L(\mathcal D_{\rm full}; \theta)$. 

\textbf{LLM Unlearning.} Let $\mathcal D_{\rm f}=\{s^1_{\rm f}, s^2_{\rm f}, ..., s^{|\mathcal D_{\rm f}|}_{\rm f}\}$ be the undesirable set that is to be unlearned from $\theta_{\rm full}$ (i.e., forget set), where $\mathcal D_{\rm f} \subset \mathcal D_{\rm full}$ and the size typically satisfies $|\mathcal D_{\rm f}|\ll|\mathcal D_{\rm full}|$, we can define the retain set as $\mathcal D_{\rm r}=\mathcal D_{\rm full}\backslash \mathcal D_{\rm f}$ to be the set of knowledge to be preserved (i.e., retain set).
Accordingly, the goal of unlearning is to derive a model $\theta_{\rm final}$ that satisfies two desiderata~\citep{tofu,muse}: (a) it \textit{forgets} the information contained in $\mathcal D_{\rm f}$, such that the model no longer provides correct answers or statements pertaining to those entities; and (b) it \textit{preserves} the knowledge in $\mathcal D_{\rm r}$, ensuring that the corresponding entities remain unaffected. {Ideally, the unlearned model should closely approximate the ground-truth model obtained by finetuning exclusively on $\mathcal D_{\rm r}$.}

\textbf{Unlearning via Task Arithmetic.} In the context of unlearning, applying task arithmetic entails computing the TV~\citep{TV} corresponding to the forget set and subsequently subtracting it from the model $\theta_{\rm full}$.
First, a forget-only finetuned model (i.e., FgtOnly model $\theta_{\rm fgt}$) is obtained on $\mathcal D_{\rm f}$ using the original pretrained model $\theta_0$ by optimizing the objective in Eq.(\ref{eq:loss}), namely, $\arg\min_{\theta \in \Theta} \mathcal L(\mathcal D_{\rm f}; \theta)$. Then the unlearned model $\theta_{\rm final}$ can simply be obtained through arithmetic operations with
\begin{equation}
    \theta_{\rm final}=\theta_{\rm full}+[-\underbrace{(\theta_{\rm fgt}-\theta_{0})}_{\text{Task Vector}}],
\end{equation}
where $\theta_0$ is used as the reference point for a purer forget-only TV (being slightly different from ~\citealp{muse} which uses $\theta_{\rm full}$). 
To address the issue of excessive forgetting on the retain set illustrated in Figure~\ref{fig:motiv}(a), an intuitive approach is to introduce a constant uniform weight $0<\omega<1$ to adjust the magnitude of the TV, i.e., $\theta_{\rm final}=\theta_{\rm full}+\omega \cdot[-(\theta_{\rm fgt}-\theta_{0})]$, thereby balancing between forgetting and retention.
However, as shown in Figure~\ref{fig:motiv}(b), since the retain gradients and the TV do not exhibit a consistent relationship across parameters, this intuitive approach may overlook divergence across parameters and is insufficient to satisfy both forgetting and retention objectives.

\begin{figure*}[t]
    \centering
    \includegraphics[width=\linewidth]{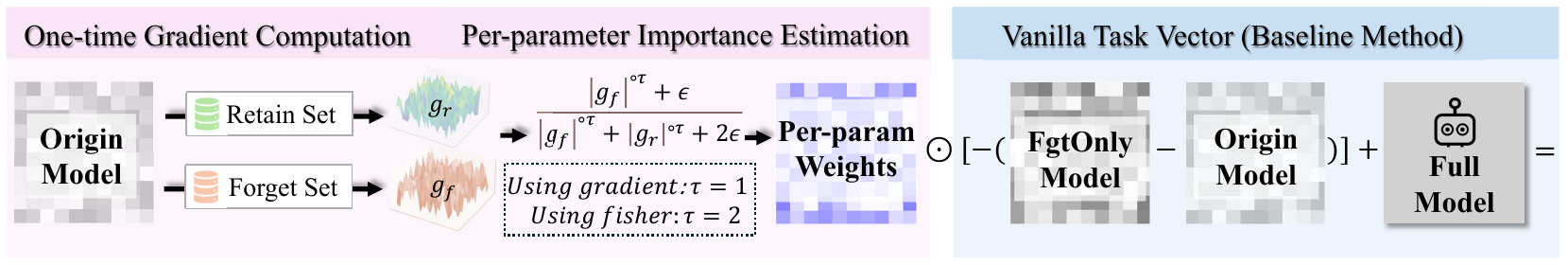}
    \vspace{-8pt}
    \caption{The framework of PerTA. PerTA rescales vanilla TV with per-parameter weights. After a one-time gradient computation on forget and retain sets, the per-parameter importance estimation introduced in Section~\ref{sec:pie} can be used to estimate the relative importance of each parameter on the forget set, either using the gradient or the Fisher information, thereby yielding the weights.}
    \label{fig:pipeline}
\end{figure*}
\textbf{Per-parameter Task Arithmetic (PerTA).}
To address these bottlenecks, we naturally propose a per-parameter weighted mechanism for TV in this work. Since each parameter contributes differently to the forget set and the retain set, we rescale TV by introducing per-parameter weights $W$, with the same dimensionality as $\theta$ (i.e., $\text{dim}(W)=\text{dim}(\theta)$). The unlearned model is therefore obtained as:
\begin{equation}\label{eq:PerTA}
    \theta_{\rm final}=\theta_{\rm full}+W\odot[-(\theta_{\rm fgt}-\theta_{0})],
\end{equation}
where $\odot$ represents per-parameter multiplication.
In $W$, larger values highlight parameters crucial for unlearning the forget set, while smaller values emphasize those important for retaining the retain set, enabling a flexible trade-off between forgetting and retention. Given the immense parameter scale of LLMs, the learning of a parametric $W$ would be prohibitively expensive. We therefore adopt a non-parametric approach to estimate $W$.
\section{Method}
\label{sec:method}

\begin{minipage}{0.4\textwidth}
The framework of PerTA and its difference from vanilla TV are shown in Algorithm~\ref{alg:pipeline} (\textcolor{violet}{violet}) and Figure~\ref{fig:pipeline}. PerTA flexibly scales TV via per-parameter multiplication between $W$ (in Eq.(\ref{eq:PerTA})) and TV. Each entry of $W$ quantifies the relative importance of its corresponding parameter for the forget set versus the retain set. 
To this end, we compute parameter gradients with respect to both the forget and retain sets (once each, with minimal overhead) and use them to construct $W$ (detailed in Section~\ref{sec:pie}). Moreover, we analyze the effectiveness of PerTA, extend it to a general form, and discuss its validity in Section~\ref{sec:common}.
\end{minipage}
\hfill
\begin{minipage}{0.57\textwidth}
\vspace{-3pt}
\begin{algorithm}[H] 
            \caption{Pipeline of PerTA}
            \begin{algorithmic}
            \STATE {\bfseries Input:} Origin/Full model $\theta_0|\theta_{\rm full}$, forget/retain set $\mathcal D_{\rm f}|\mathcal D_{\rm r}$, hyperparameter $E, \alpha$
            \STATE {\bfseries Output:} Unlearned model $\theta_{\rm final}$
            \STATE \textcolor{gray}{\# Step 1: Calculting $\theta_{\rm fgt}$ required by TV}
            \STATE $\theta_{\rm fgt} \leftarrow \theta_0$ 
                    \FOR{$e=1, \dots, E$}
                        \STATE $\theta_{\rm fgt} \leftarrow \theta_{\rm fgt} - \alpha \nabla\mathcal L(\mathcal D_{\rm f}; \theta_{\rm fgt})$
                    \ENDFOR
            \STATE \textcolor{gray}{\# Step 2.1 One-time gradient computation}
            \STATE \textcolor{violet}{$g_{\rm f} \leftarrow \nabla \mathcal L(\mathcal D_{\rm f};\theta_0)$, $g_{\rm r} \leftarrow \nabla \mathcal L(\mathcal D_{\rm r};\theta_0)$}
            \STATE \textcolor{gray}{\# Step 2.2 Per-parameter importance estimation}
            \STATE \textcolor{violet}{$W \leftarrow\frac{|g_{\rm f}|^\tau+\epsilon}{|g_{\rm f}|^{\tau}+|g_{\rm r}|^\tau+2\epsilon}$} (using Eq.(\ref{eq:grad}) or Eq.(\ref{eq:fisher}))
            \STATE \textcolor{gray}{\# Step 3: Task arithmetic
            \STATE $\theta_{\rm final}\leftarrow\theta_{\rm full}+\textcolor{violet}{W\odot}[-(\theta_{\rm fgt}-\theta_{0})]$ (using Eq.(\ref{eq:PerTA}))}
            \end{algorithmic}
            \label{alg:pipeline}
        \end{algorithm}
\end{minipage}

\subsection{Per-parameter Importance Estimation}
Let $W=[w_1, w_2, ..., w_n]$ be the scaling weights corresponding to the model parameters $\theta=[q_1, q_2, ..., 
q_n]$ with $n$ parameters. Each weight satisfies $w_i\in [0, 1], 1\leq i \leq n$. Values $w_i$ closer to 1 indicate that TV at $q_i$ should be kept, while values approaching 0 downweight TV at $q_i$. 

\label{sec:pie}

\textbf{Using Absolute Gradient (PerTA-grad).}
Since the importance is independent of gradient direction, the absolute magnitude of the parameter gradients~\citep{gradient1,gradient2} provides a natural measure of importance. {While gradient estimation using either $\theta_0$ or $\theta_{\rm full}$ is justifiable, we adopt $\theta_0$ here because $\theta_0$ is a more neutral initialization model that does not contain training data from the forget or the retain set. However, in practice, estimating gradients on $\theta_0$ or on $\theta_{\rm full}$ makes a negligible difference (see Appendix~\ref{app:model} for a detailed discussion).} Let $\nabla \mathcal L(\mathcal D_{\rm f};\theta_0),\nabla \mathcal L(\mathcal D_{\rm r};\theta_0)$ be the gradients of forget and retain sets. The weight for each parameter can be computed as the relative contribution of the forget set gradient to the total gradient magnitude, where $W$ can be formulated as:
\begin{equation}\label{eq:grad}
    W_{\rm grad}=\frac{|\nabla \mathcal L(\mathcal D_{\rm f};\theta_0)|+\epsilon}{ |\nabla \mathcal L(\mathcal D_{\rm r};\theta_0)|+ |\nabla \mathcal L (\mathcal D_{\rm f};\theta_0)|+2\epsilon},
\end{equation}
where $\epsilon$ is a small constant to avoid division by zero. 
Substituting Eq.(\ref{eq:grad}) into Eq.(\ref{eq:PerTA}) yields the final unlearned model. $W_{\rm grad}$ treats all deviations linearly. Next, we also propose a non-linear alternative.

\textbf{Using Diagonal Fisher Information Approximation (PerTA-fisher).} 
The diagonal of the Fisher Information Matrix~\citep{fisher1,fisher2} is widely used to reflect the sensitivity of parameters to the data. The computation of its diagonal entries can be simplified as the squared gradients (see Appendix~\ref{app:fisher} for a detailed proof). Accordingly, $W$ can also be expressed as:
\begin{equation}\label{eq:fisher}
    W_{\rm fisher}=\frac{\nabla \mathcal L^2(\mathcal D_{\rm f};\theta_0)+\epsilon}{\nabla \mathcal L^2(\mathcal D_{\rm r};\theta_0)+ \nabla \mathcal L^2(\mathcal D_{\rm f};\theta_0)+2\epsilon}.
\end{equation}
Similar to $W_{\rm grad}$, substituting Eq.(\ref{eq:fisher}) into Eq.(\ref{eq:PerTA}) yields the final unlearned model, as detailed in Algorithm~\ref{alg:pipeline}. Both $W_{\rm grad}$ and $W_{\rm fisher}$ essentially estimate the per-parameter importance of the forget set by computing the relative magnitude of gradients on $\mathcal D_{\rm f}$. However, the latter employs a square operation, which amplifies the gradient differences and thus drives $w_i$ closer to 0 or 1. The detailed discussion is in the next subsection.

\subsection{Discussion and A General Form}
\label{sec:common}
\textbf{Discussion about PerTA.} Denoting $g_{\rm f}\triangleq \nabla \mathcal L(\mathcal D_{\rm f};\theta_0) $, $g_{\rm r}\triangleq  \nabla \mathcal L(\mathcal D_{\rm r};\theta_0)$, we further explore the effectiveness of $W_{\rm grad}$ and $W_{\rm fisher}$, and their difference.
\begin{definition}
    (Retain-forget ratio). \textit{
    For a parameter $q_i$ in an LLM, the retain-forget ratio reflects its relative importance for retention versus forgetting.
    Denoting $[g_{\rm r}]_i$ and $[g_{\rm f}]_i$ to be the gradients of $q_i$ on the retain and forget sets, the retain-forget ratio can be represented as $$r_i=({|[g_{\rm r}]_i|+\epsilon})/({|[g_{\rm f}]_i|+\epsilon}).$$ 
    When $r_i \geq 1$, the retain set dominates for $q_i$, and the forget set dominates when $r_i < 1$.}
\end{definition}
Hence, we obtain the following proposition.
\begin{proposition}\label{prop:diff}
    For parameter $p_i$, we denote its corresponding weights calculated with PerTA-grad, PerTA-fisher to be $\omega_i^{\rm grad}$ and $\omega_i^{\rm fisher}$ respectively. Then we have:
    \begin{align}
        \frac{1}{2} \geq \omega_i^{\rm grad}\geq \omega_i^{\rm fisher}\geq 0, \text{    when  } r_i \geq 1 \text{;   }
        \frac{1}{2} < \omega_i^{\rm grad}< \omega_i^{\rm fisher}< 1, \text{    when   } r_i < 1, \notag
    \end{align}
    which is proved in Appendix~\ref{app:grad_fisher}. 
\end{proposition}
This implies that when $p_i$ has a larger influence on the retain set, PerTA applies a smaller reweighting to TV in order to reduce forgetting, and vice versa. Notably, compared to PerTA-grad, PerTA-fisher yields weights that are closer to 0 or 1, thereby creating a cleaner separation between parameters to be edited and to be preserved.

\textbf{A General Form.} Besides, the determination of $W$ is not limited to the aforementioned approaches. Here, we express $W$ in a more general form--as a function of $g_{\rm f}$ and $g_{\rm r}$:
\begin{equation}
    W_{\rm general}=f_{\rm oprt}(g_{\rm f}, g_{\rm r}), \notag
\end{equation}
where $f_{\rm oprt}(\cdot,\cdot)$ is a custom operation. Then both $W_{\rm grad}$ and $W_{\rm fisher}$ can be represented with $f_{\rm oprt}(A, B)=|A|^{\circ \tau}/(|A|^{\circ \tau}+|B|^{\circ \tau})$, where $\circ\tau$ is the per-parameter $\tau$-th power and the case $\tau=1$ and $\tau=2$ correspond to $W_{\rm grad}$ and $W_{\rm fisher}$, respectively. 

\textbf{Discussion about $f_{\rm oprt}(\cdot, \cdot)$.} In addition to the absolute gradient and diagonal Fisher Information approximation we applied, other operations---such as the SoftMax-based formulation $f_{\rm oprt}(A, B)=\exp(|A|)/(\exp(|A|)+\exp(|B|))$---can also be employed (see Section~\ref{sec:exp_abl} for detailed results and discussions). Moreover, $W_{\rm general}$ subsumes more general cases: when $f_{\rm oprt}(A, B)=1$, it degenerates to vanilla TV, whereas when $f_{\rm oprt}(A, B)=w$, PerTA employs the uniform constant $w$ to balance forgetting and retaining. Denoting weight $w_i$ for parameter $q_i$ as $w_i=[f_{\rm oprt}(g_{\rm f}, g_{\rm r})]_{i}$ and the corresponding gradients are $[g_{\rm f}]_i$ and $[g_{\rm r}]_i$, in the following, we discuss the design of $f_{\rm oprt}(\cdot, \cdot)$:
\begin{itemize}
    \item {Intuitively, it should satisfy $[f_{\rm oprt}(g_{\rm f}, g_{\rm r})]_i\rightarrow 1$ when $|[g_{\rm f}]_i| \gg |[g_{\rm r}]_i|$, and $[f_{\rm oprt}(g_{\rm f}, g_{\rm r})]_i\rightarrow 0$ when $|[g_{\rm r}]_i| \gg |[g_{\rm f}]_i|$. This is because TV is the vector for forget set $\mathcal D_{\rm f}$: when $|[g_{\rm f}]_i|$ is large, the parameter $q_i$ is crucial for unlearning, and thus the rescaled TV should preserve its value; conversely, when $|[g_{\rm r}]_i|$ is large, the parameter is critical for retention, and the TV should therefore be scaled down. $W_{\rm grad}$ and $W_{\rm fisher}$ are consistent with this intuition (see Appendix~\ref{app:lim}).}
    \item Empirically, we explored several straightforward ways of designing $f_{\rm oprt}(\cdot, \cdot)$ and found that $W_{\rm grad}$ and $W_{\rm fisher}$ in this paper perform best among them, as detailed in the ablation studies from Section~\ref{sec:exp_abl}. Naturally, the choice of $f_{\rm oprt}(\cdot, \cdot)$ is not unique, and we hope our work will inspire further exploration and discussion.
\end{itemize}

\section{Experiments}
\label{sec:exp}
\textbf{Baselines and Benchmarks.}
Experiments are conducted on the widely used unlearning benchmark TOFU~\citep{tofu} (covering three tasks with 1\%, 5\%, and 10\% of the data unlearned) and on MUSE News~\citep{muse}. On TOFU, following \citealp{openunlearning}, we employ Llama-3.2 1B and 3B Instruct models~\citep{llama} and evaluate them using five metrics: (1) Forget Quality (FQ)~\citep{tofu}, which measures the effectiveness of unlearning (higher is better, we use log transformation in this paper); (2) Model Utility (MU)~\citep{tofu}, which quantifies the model’s usefulness in retaining original knowledge (higher is better); (3) Extraction Strength (ES)~\citep{es} of the forget set, defined as the proportion of repeated content start positions in the forget set (lower is better); (4) ES of the retain set, defined analogously on the retain set (higher is better); (5) Gibberish (Gib), which represents the probability---determined by a binary classifier~\citep{gibberish-detector-2021}---that answers to forget-set queries are non-gibberish (higher is better) and (6) ROUGE-L (ROUGE)~\citep{rouge}, the proportion of the longest common sub-sequence between the ground truth and the answers. Additional dataset-related information is provided in Appendix~\ref{app:data}, while detailed definitions of the metrics are given in Appendix~\ref{app:matrix}.

As for the baselines, for training-based methods we evaluate the mainstream approaches GA~\citep{yao2024large}, GD~\citep{GD}, NPO~\citep{NPO}, and NPO+ (NPO combined with GD). For task-arithmetic methods, we test vanilla TV~\citep{TV} and our proposed method. In addition, we report the metrics of the full model before unlearning, alongside those of a ground-truth model trained solely on the retain set~\citep{tofu}, as references. Detailed information about the baselines and implementation can be found in Appendices~\ref{app:baseline} and \ref{app:imp}, respectively.

\textbf{Performance Comparison.}
The results of FQ and MU on the three TOFU tasks with 1\%, 5\%, and 10\% samples to be unlearned are shown in Figure~\ref{fig:res_mufq} (see more metrics in Appendix~\ref{app:graph}). The ground-truth results are shown as black pentagram markers. The FQ metric measures the p-value of distributional differences from the ground truth; we perform the logarithmic transformation to better highlight variations. Dark-blue and purple circles denote methods PerTA-grad and PerTA-fisher, respectively. On simpler tasks (e.g., unlearning 1\% of the data), most training-based methods maintain model utility, while the task-arithmetic method TV achieves higher FQ but at the cost of MU. Our PerTA-grad and PerTA-fisher improve MU relative to TV and yield results closer to the ground truth. On more challenging tasks (e.g., unlearning 5\% or 10\%), training-based methods degrade: MU for GA and NPO drops to nearly zero, and their FQ becomes both lower and unstable (with larger variance). In contrast, our PerTA-grad and PerTA-fisher consistently outperform both training-based and task-arithmetic baselines in FQ and MU, confirming the effectiveness of PerTA in achieving unlearning while preserving model capability.

To examine why PerTA outperforms TV among task arithmetic methods, we evaluate four dimensions: \textit{forget}, \textit{retain}, \textit{real}, and \textit{facts}~\citep{tofu}, corresponding to the forget set, retain set, original authors, and world facts. The first two measure forgetting and retention of post-training knowledge, while the latter two assess preservation of pretrained knowledge. ROUGE is used to measure similarity to reference answers. As shown in Figure~\ref{fig:res_rouge} (see Appendix~\ref{app:graph} for more results), TV performs well on real authors and world facts, and PerTA preserves this capability. However, for post-training knowledge, TV suffers from over-forgetting, whereas PerTA significantly narrows the gap to the ground truth. In more challenging settings (e.g., unlearning 5\% and 10\%), TV falls far below the ground truth, while PerTA nearly doubles TV’s performance, being much closer to the reference.
\begin{figure*}[t]
    \centering
    \includegraphics[width=\linewidth]{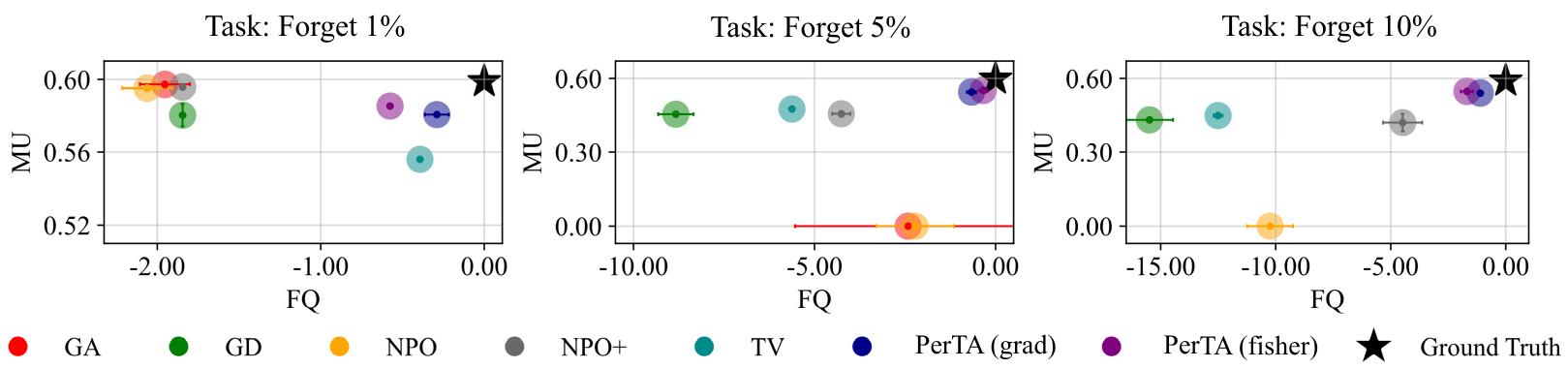}
    \vspace{-8pt}
    \caption{MU and FQ results of different methods on TOFU (using Llama-3.2 1B Instruct), where circle markers denote values and horizontal and vertical bars at circle centers represent error bars. }
    \label{fig:res_mufq}
\end{figure*}
\begin{figure*}[t]
    \centering
    \includegraphics[width=\linewidth]{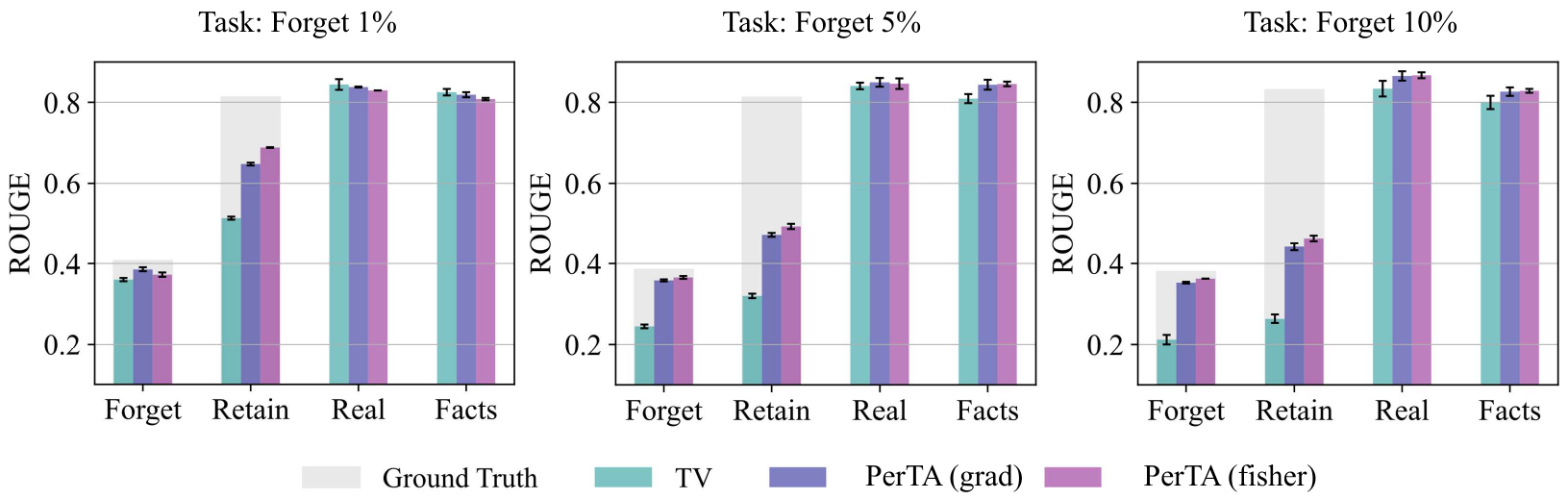}
    \vspace{-8pt}
    \caption{Four-dimension ROUGE results of task arithmetic-based methods on TOFU (using Llama-3.2 1B Instruct). Ground-truth results on forget and retain sets are marked with a gray background.}
    \label{fig:res_rouge}
\end{figure*}

\begin{table*}[t]
\caption{Average results of different methods on three tasks (unlearning 1\%, 5\%, 10\% of TOFU). The references are in \textcolor{gray}{gray font}, the best two are in \textbf{bold}, and ours are \textcolor{black}{\colorbox{gray!30}{highlighted}}. `Full' and `GT' represent the model before unlearning and the ground truth model.}
\centering
\vspace{0.1cm}
\begin{small}
\resizebox{\textwidth}{!}{

\begin{tabular}{c|ccccc|ccccc}
\toprule
              & FQ↑                            & MU↑                          & ES($\mathcal D_{\rm f}$)↓                    & ES($\mathcal D_{\rm r}$)↑                   & Gib↑                   & FQ↑                            & MU↑                          & ES($\mathcal D_{\rm f}$)↓                    & ES($\mathcal D_{\rm r}$)↑                                             & Gib↑                   \\
\midrule
Model Size    & \multicolumn{5}{c|}{1B}                                                                                                                                                                    & \multicolumn{5}{c}{3B}                                                                                                 \\
\midrule
Full            & \textcolor{gray!50}{ -11.808} & \textcolor{gray!50}{ 0.599} & \textcolor{gray!50}{ 0.726} & \textcolor{gray!50}{ 0.737} & \textcolor{gray!50}{ 0.871} & \textcolor{gray!50}{ -13.960} & \textcolor{gray!50}{ 0.666} & \textcolor{gray!50}{ 0.899} & \textcolor{gray!50}{ 0.884} & \textcolor{gray!50}{ 0.868} \\
GT        & \textcolor{gray!50}{ 0.000}   & \textcolor{gray!50}{ 0.596} & \textcolor{gray!50}{ 0.064} & \textcolor{gray!50}{ 0.748} & \textcolor{gray!50}{ 0.894} & \textcolor{gray!50}{ 0.000}   & \textcolor{gray!50}{ 0.657} & \textcolor{gray!50}{ 0.066} & \textcolor{gray!50}{ 0.887} & \textcolor{gray!50}{ 0.887} \\

\midrule
\multicolumn{11}{c}{Training-based}                                                                                                                                                                               \\
\midrule
GA                  & -81.114                        & 0.199                        & 0.086                        & 0.244                        & 0.484                        & -81.257                        & 0.383                        & 0.125                        & 0.331                        & 0.593                        \\
GD                  & -8.720                         & 0.491                        & 0.112                        & 0.295                        & 0.789                        & -13.959                        & 0.589                        & 0.192                        & 0.437                        & 0.677                        \\
NPO                 & -4.842                         & 0.198                        & 0.086                        & 0.246                        & 0.592                        & -4.508                         & 0.380                        & 0.123                        & 0.334                        & 0.637                        \\
NPO+                & -3.528                         & 0.493                        & 0.122                        & 0.316                        & 0.911                        & -5.413                         & 0.587                        & 0.147                        & 0.415                        & 0.898                        \\

\midrule
\multicolumn{11}{c}{Arithmetic-based}                                                                                                                                                                                                                                                                                                                                                            \\
\midrule   
TV  & -6.174                         & 0.495                        & \textbf{0.059}               & 0.207                        & \textbf{0.914}               & -5.284                         & 0.612                        & \textbf{0.058}               & 0.304                        & \textbf{0.921}               \\

\rowcolor{gray!30} PerTA-grad  & \textbf{-0.686}                & \textbf{0.556}               & \textbf{0.072}               & \textbf{0.376}               & \textbf{0.915}               & \textbf{-0.669}                & \textbf{0.664}               & \textbf{0.082}               & \textbf{0.563}               & \textbf{0.913}               \\

\rowcolor{gray!30} PerTA-fisher   & \textbf{-0.867}                & \textbf{0.562}               & 0.080                        & \textbf{0.414}               & 0.908                        & \textbf{-1.211}                & \textbf{0.665}               & 0.092                        & \textbf{0.613}               & 0.895                       
\\            
\bottomrule
\end{tabular}}

\end{small}
\label{tab:res}
\end{table*}

\label{sec:exp_abl}
\begin{figure}[t]
    \centering
    \includegraphics[width=0.96\linewidth]{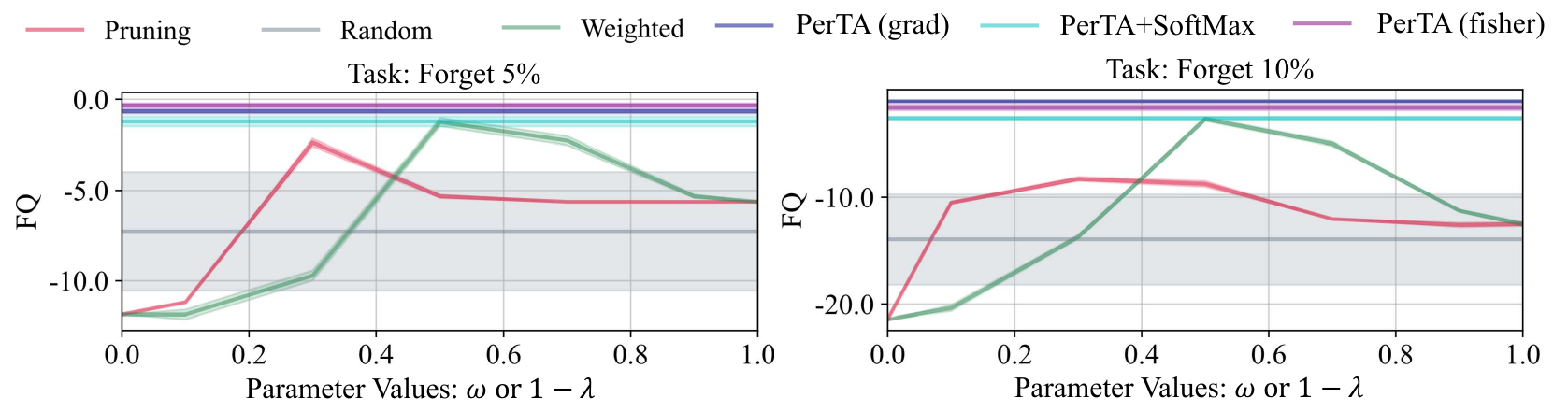}
    \vspace{-3pt}
    \caption{Results of FQ (↑) using different $f_{\rm oprt}$ on two challenging tasks (unlearning 5\% and 10\% of TOFU, Llama-3.2 1B Instruct). The shaded region indicates the error bounds.}
    \vspace{-8pt}
    \label{fig:ablation}
\end{figure}

\textbf{More Backbones and Benchmarks.} Table~\ref{tab:res} reports the average results of different methods across the three unlearning tasks, with both 1B and 3B model sizes considered to examine the effect of different LLM backbones (see complete results in Appendix~\ref{app:quant}). The results show that the baseline TV, compared with training-based methods, suffers from excessive forgetting on the retain set (low ES ($\mathcal D_{\rm r}$)), while our PerTA substantially improves ES ($\mathcal D_{\rm r}$) without significantly reducing ES ($\mathcal D_{\rm f}$). Moreover, PerTA delivers notable gains in FQ (e.g., with the ground truth being $0$, TV achieves $-6.174$ and $-5.284$, whereas PerTA-grad reaches about $-0.67$ and PerTA-fisher about $-1$) and MU (e.g., on the 1B model, PerTA raises MU from $0.495$ to $0.556$ by PerTA-grad or $0.562$ by PerTA-fisher, narrowing the gap to the ground truth to within $0.04$). On larger backbones such as 3B, PerTA maintains improvements in both FQ and MU while further increasing ES ($\mathcal D_{\rm r}$) without compromising ES ($\mathcal D_{\rm f}$). These results demonstrate the effectiveness of PerTA in achieving unlearning while preserving utility across different model scales. Additionally, results in Appendix~\ref{app:quant} show that PerTA is also effective on MUSE.

\textbf{Ablation (General Form) Studies.} 
Figure~\ref{fig:ablation} shows the curves of FQ when different $f_{\rm oprt}(\cdot, \cdot)$ are selected under varying hyperparameters. In addition to PerTA-grad and PerTA-fisher proposed in Eq.(\ref{eq:grad}) and Eq.(\ref{eq:fisher}), we consider several straightforward designs: (1) `Pruning': removing (i.e., $f_{\rm oprt}(A, B)=0$) the $\lambda\%$ smallest weights in TV to mitigate over-forgetting and maintain others (i.e., $f_{\rm oprt}(A, B)=1$), where $\lambda=0$ reduces to vanilla TV; (2) `Random': setting weights in $W$ to random values uniformly sampled between 0 and 1 with $f_{\rm oprt}(A, B)=\text{rand([0, 1])}$; (3) `Weighted': using a constant $\omega$ to rescale TV with $f_{\rm oprt}(A, B)=\omega$, where $\omega=1$ reduces to vanilla TV; and (4) `SoftMax': determining $f_{\rm oprt}(A, B)=\exp(|A|)/(\exp(|A|)+\exp(|B|))$ in the SoftMax form. 

Among these, `Pruning' and `Weighted' methods vary with $\lambda$ or $\omega$, as shown in Figure~\ref{fig:ablation}. We observe that `Pruning' performs poorly on more challenging tasks (e.g., unlearning 10\%), `Random' exhibits very high variance, and `Weighted' can achieve reasonable results when the optimal constant $\omega$ is chosen but is highly sensitive to the hyperparameter. The `SoftMax' method represents a successful design of $f_{\rm oprt}(\cdot, \cdot)$, yet our PerTA-grad and PerTA-fisher still outperform other possible designs.

\begin{wrapfigure}{r}{0.55\textwidth}    \centering
    \includegraphics[width=\linewidth]{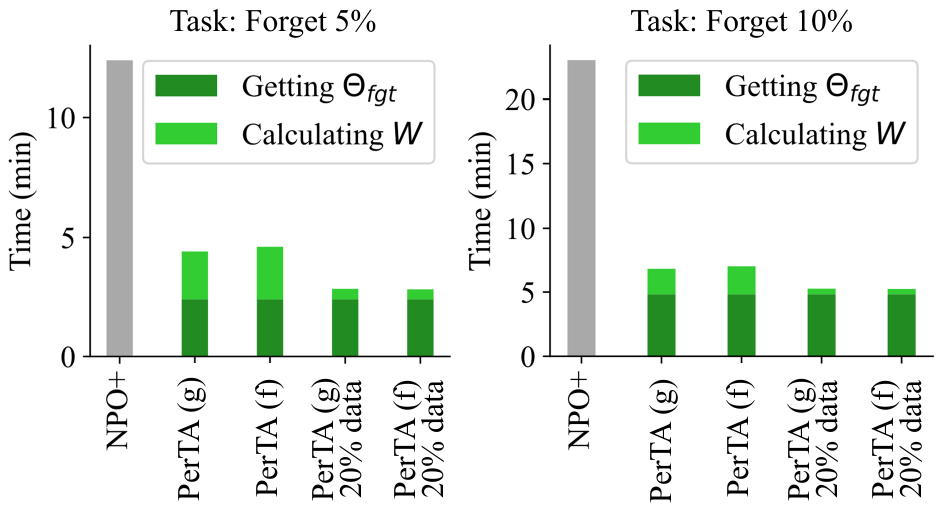}
    \caption{Time comparison of the best-performing training-based method NPO+ and our PerTA (unlearning 5\% and 10\%, Llama-3.2 1B Instruct).}
    \label{fig:time}
\end{wrapfigure}

\textbf{Time Efficiency Discussion.}
Figure~\ref{fig:time} shows the runtime comparison between the best-performing training-based method, NPO+, and our PerTA. 
Unlike training-based approaches that require repeated iterations, the runtime of PerTA can be decomposed into: the time to obtain $\theta_{\rm fgt}$, the time to compute $W$, and the time for task arithmetic, where the latter is negligible. It can be observed that PerTA inherits the advantage of task arithmetic---significantly reducing runtime---and this advantage becomes more pronounced as task complexity increases (i.e., when unlearning larger proportions). Moreover, as shown previously, estimating gradients with only 20\% of the data already yields competitive results, suggesting that runtime can be further reduced. Together, these findings highlight the time efficiency of PerTA. More results are in Appendix~\ref{app:quant}.

\begin{figure}[t]
    \centering
    \includegraphics[width=\linewidth]{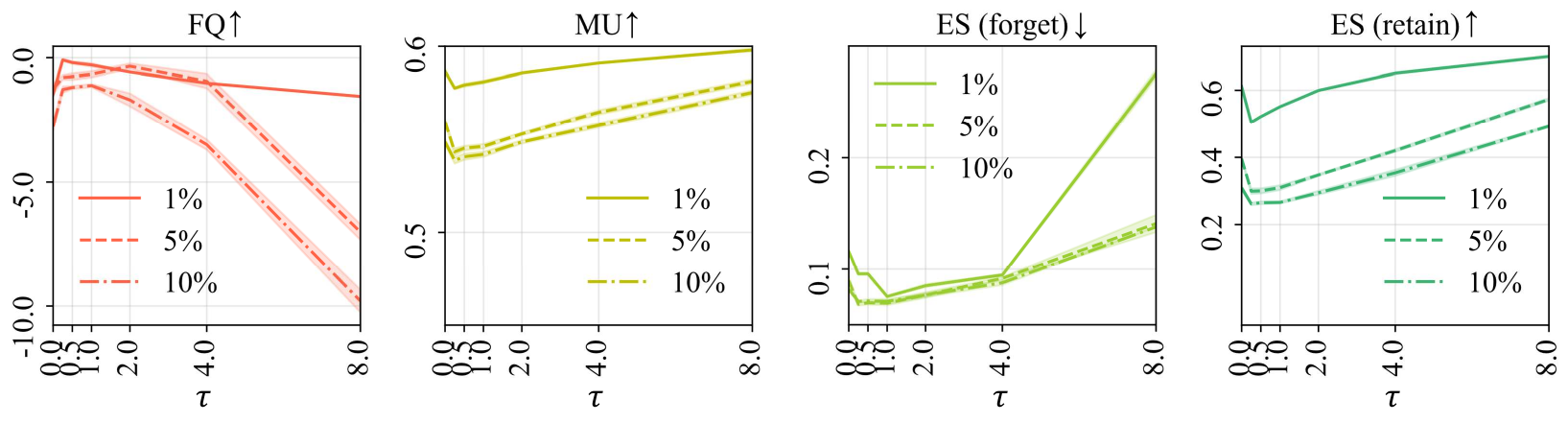}
    \vspace{-8pt}
    \caption{Results of alternative variants in $f_{\rm oprt}(A, B)=|A|^{\circ \tau}/(|A|^{\circ \tau}+|B|^{\circ \tau})$ with different $\tau$s (Llama-3.2 1B Instruct). 1\%, 5\%, 10\% tasks are distinguished with different line types.}
    \label{fig:tau}
\end{figure}
\begin{figure}[t]
    \centering
    \includegraphics[width=\linewidth]{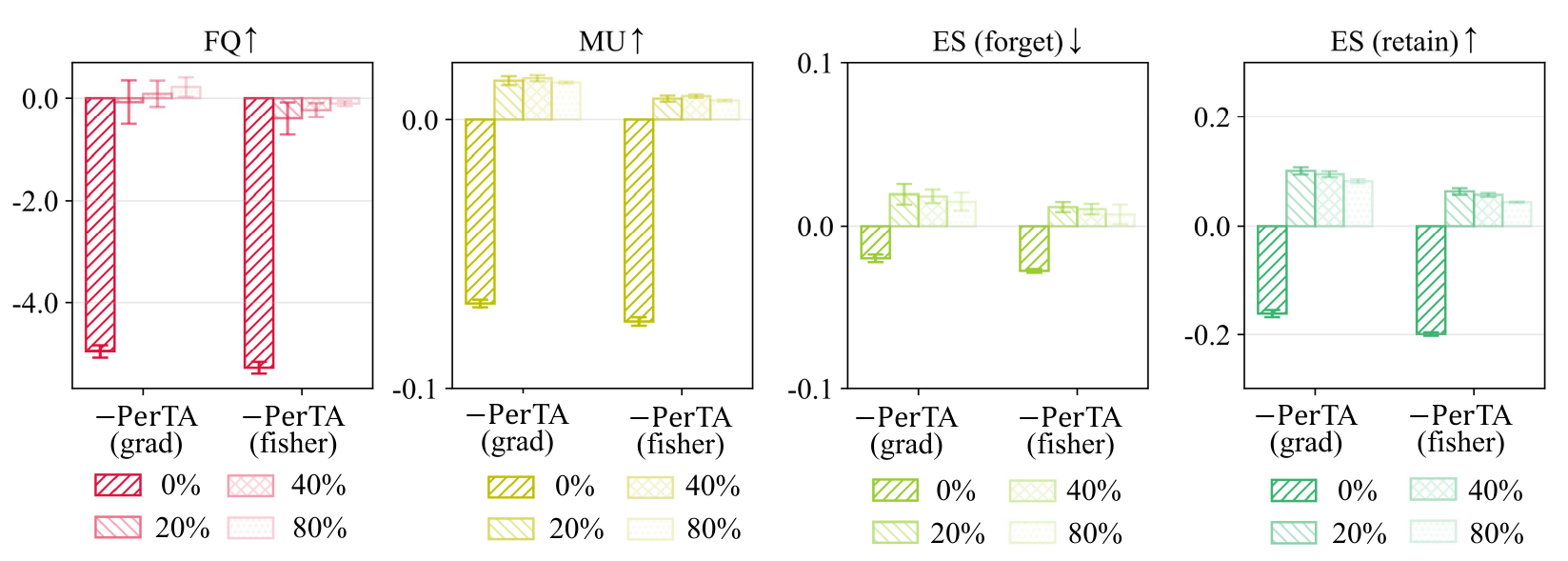}
    \vspace{-8pt}
    \caption{Residual results of the metrics when using only 20\%, 40\%, and 80\% of the samples compared to using the full set (unlearning 5\%, Llama-3.2 1B Instruct). 0\% denotes vanilla TV.}
    \label{fig:parts}
\end{figure}

\textbf{Alternative Variants Analysis.}
When retaining the form of $f_{\rm oprt}(A, B)=|A|^{\circ \tau}/(|A|^{\circ \tau}+|B|^{\circ \tau})$ but not using PerTA-grad and PerTA-fisher, different $\tau$s can be applied. We conduct experiments for $\tau \in \{0,0.25,0.5,1,2,4,8\}$, with results shown in Figure~\ref{fig:tau} and Appendix~\ref{app:quant}. The cases of $\tau=1,2$ correspond to our PerTA-grad and PerTA-fisher, respectively. Our methods strike a balance between forgetting and retaining: among the different $\tau$-based variants, they achieve relatively strong FQ and ES ($\mathcal D_{\rm f}$) while keeping MU and ES ($\mathcal D_{\rm r}$) at a reasonable level.

\textbf{Sample Efficiency Discussion.}
In this experiment, we estimate $W_{\rm grad}$ and $W_{\rm fisher}$ using 0\%, 20\%, 40\%, and 80\% of the total samples, where 0\% corresponds to vanilla TV and the other three represent PerTA with reduced sample sizes. The differences in metrics compared to using the full dataset are shown in Figure~\ref{fig:parts} and Appendix~\ref{app:quant}. It is observed that using only one-fifth of the samples already yields results comparable to those obtained with the full dataset, and better than vanilla TV. This demonstrates that PerTA is also sample-efficient, which can be beneficial in further reducing the unlearning time.

\textbf{More Experiments.}
The experimental results comparing gradient $g_{\rm f}, g_{\rm r}$ prediction using $\theta_0$ or $\theta_{\rm full}$ are provided in Appendix~\ref{app:model}. Visualizations and discussions of the magnitude of TV and $W$ across different attention layers of the LLM are presented in Appendix~\ref{app:visual}. {Results on larger or alternative LLM models are in Appendix~\ref{app:other}. Results under quantization attacks are detailed in Appendix~\ref{app:attack}.}

\section{Conclusion}
To address the issue of potentially over-forgetting on the retain set when using vanilla TV, we proposed PerTA to rescale TV, where the weight matrix is estimated using absolute gradients or the diagonal Fisher Information approximation. The effectiveness of PerTA is validated by both theoretical analysis and empirical evidence.
\clearpage
\bibliography{PerTA}
\bibliographystyle{plain}

\clearpage
\appendix
\section{Appendix 1: More Training Information}
\subsection{Dataset Information}
\textbf{TOFU.} The TOFU dataset\footnote{\url{https://huggingface.co/datasets/locuslab/TOFU}} is designed as a benchmark to assess how well LLMs can perform unlearning on practical tasks. It contains 4000 question-answer pairs derived from autobiographies of 200 entirely fictional authors, all generated by GPT-4. The task involves evaluating a finetuned model’s ability to unlearn when exposed to different proportions (i.e., unlearning 1\%, 5\%, 10\%) of the forget set.

\textbf{MUSE.} MUSE is a benchmark designed to evaluate machine unlearning. It centers on two major forms of textual content where unlearning is often necessary: news reports (News) and literary works (Books). The MUSE-News subset\footnote{\url{https://huggingface.co/datasets/muse-bench/MUSE-News}} specifically includes BBC articles published after August 2023.
\label{app:data}
\subsection{Metric Discussion}
\label{app:matrix}
{In fact, the choice of evaluation metrics for unlearning has long been an active and debated research topic. Assessing unlearning performance typically requires considering multiple aspects and dimensions. In this paper, we adopt the metrics used in \citep{tofu}, which are also widely employed by mainstream methods such as \citep{openunlearning,GRU,satimp}}.

Similar to Section~\ref{sec:pre}, we define the new knowledge dataset to post-train the LLM as $\mathcal D=\{s^1,s^2, ..., s^{|\mathcal D|}\}$ consisting of $|\mathcal D|$ sequences, where each sequence $s=[t_1, t_2, ..., t_{|s|}]$ contains $|s|$ tokens. To split $s$ into questions and answers, we can also write $s=[x, y]$. Then the probability of $y$ given $x$ is defined as $$\text P(\mathcal D;\theta)=\mathbb E_{[x, y] \sim \mathcal D} p(y|x; \theta)^{\frac{1}{|y|}}=\mathbb E_{[x, y] \sim \mathcal D}[\prod^{|y|}_{i=1}p(y_i|[x, y_{<i}];\theta)]^\frac{1}{|y|},$$ which is normalize for answer length as a common practice~\citep{cho2014properties}. Denoting $\mathcal Y_{\rm pret}$ as the set of incorrect answers with the same template as $y$, the truth ratio can be defined as $$\text{Tr}(\mathcal D; \theta)=\mathbb E_{[x,y] \sim \mathcal D}\frac{\frac{1}{|\mathcal Y_{\rm pret}|}\sum\nolimits_{\tilde y \in \mathcal Y_{\rm pret}} \text P(\tilde y|x)}{\text P(y|x)}.$$ Besides, by obtaining $\arg\max_{t_i} p(t_i|t_{<i};\theta)$, the output texts of LLM given prompt $t_{<i}=[t_1, ..., t_{i-1}]$ is defined as $f(t_{<i}; \theta)$.

\textbf{ROUGE-L (ROUGE).} Denoting the length of the longest common sub-sequence considering string $a$ and $b$ as $\text{LCS}(a, b)$, then the ROUGE-L metric can be defined for model $\theta$ and dataset $\mathcal D$ as $$\text{ROUGE}(\mathcal D;\theta)=\mathbb E_{[x, y] \sim \mathcal D} \frac{\text{LCS}(y, f(x;\theta))}{|y|}.$$ The bigger ROUGE-L is, the more similar the references and output answers of LLM are.

\textbf{Extraction Strength (ES).} ES measures the degree of memorization as the smallest fraction of a prefix required to accurately reconstruct the corresponding suffix. It can be formulated as $$\text{ES}(\mathcal D;\theta)=\mathbb E_{[x,y] \sim \mathcal D} [1-\frac{1}{|y|}\min_k\{k|f([x, y_{<k}];\theta)=y_{>k}\}].$$

\textbf{Forget Quality (FQ).} {The goal of unlearning is for the final model to approximate the model trained on the retain set only.} Therefore, FQ is used to assess unlearning by statistically comparing the truth ratio $\text{Tr}(y|x; \theta)$ distributions of the unlearned model $\theta$ and the model $\theta_{\rm retain}$ trained on retain set only with KS-Test~\citep{ks-test}, producing higher scores when the two distributions are closely aligned: $$\text{FQ}(\mathcal D_{\rm f};\theta)=\text{KS}(\text{Tr}_{[x, y] \sim \mathcal D_{\rm f}}(y|x; \theta), \text{Tr}_{[x, y] \sim \mathcal D_{\rm f}}(y|x; \theta_{\rm retain})),$$
where $\text{KS}(\cdot, \cdot)$ is the KS-Test function and $\mathcal D_{\rm f}$ is the forget set.

\textbf{Model Utility (MU).} MU measures how well a model performs after unlearning, on both the retain set and general knowledge. It is defined as the harmonic mean of three metrics--probability, ROUGE, and Truth Ratio--evaluated across three levels: retain set $\mathcal D_{\rm r}$, real authors $\mathcal D_{\rm a}$, and world factual knowledge $\mathcal D_{\rm w}$: $$\text{MU}(\theta)=\frac{1}{\sum\nolimits_{\mathcal D \in \{ \mathcal D_{\rm f}, \mathcal D_{\rm a}, \mathcal D_{\rm w}\}}[\frac{1}{\text P(\mathcal D;\theta)}+\frac{1}{\text{Tr}(\mathcal D;\theta)}+\frac{1}{\text{ROUGE}(\mathcal D;\theta)}]}.$$ Different from the retain set, when calculating the probability on $\mathcal D_{\rm a}$ and $\mathcal D_{\rm r}$, function $\text P$ is defined as $\text{P}(x|y; \theta)=p(y|x;\theta)/\sum\nolimits_{\tilde y \in \mathcal Y_{\rm choice}}p(\tilde y|x;\theta)$, where $\mathcal Y_{\rm choice}$ is the given possible answer set.

\textbf{Gibberish (Gib).} Unlearning can negatively impact model fluency, especially on the forget set, leading to incoherent or meaningless outputs. To measure this phenomenon, a classifier-based score\footnote{\url{https://huggingface.co/madhurjindal/autonlp-Gibberish-Detector-492513457}} is employed to determine whether the generated text resembles gibberish.

\subsection{Training-based Methods}
\label{app:baseline}
Training-based approaches generally employ a specifically designed loss function to facilitate unlearning in LLMs. The training procedure involves iteratively computing this loss and updating the model's weights. After a number of iterations, the process concludes, resulting in the final model. This section details the loss functions used in the training-based methods discussed in this work.

\textbf{GA.} GA is the pioneering work that first maximize the loss of the forget set. As the general loss function of LLM learning $\mathcal L(\mathcal D;\theta)$ is defined in Eq.(\ref{eq:loss}), the loss of GA can be formulated as $$\mathcal{L}_{\rm GA}(\theta)=-\mathcal L(\mathcal D_{\rm f};\theta).$$

\textbf{GD.} To avoid over-forgetting the retain set, GD performs gradient descent on the retain set: $$\mathcal{L}_{\rm GD}(\theta)=-\mathcal L(\mathcal D_{\rm f};\theta)+\alpha L(\mathcal D_{\rm r};\theta),$$ where $\alpha$ is the coefficient to balance between unlearning and retention.

\textbf{NPO.} NPO constructs its loss function inspired by the dis-preferred component of DPO. This type of loss is suitable for the question-answer pairs. Thus, the loss function is $$\mathcal L_{\rm NPO}(\theta)=-\frac{2}{\beta}\mathbb E_{[x,y] \sim \mathcal D_{\rm f} }\log \sigma (- \beta \log(\frac{p(y|x;\theta)}{p(y|x;\theta_{\rm full})})),$$ where $\sigma(\cdot)$ represents the Sigmoid function and $\beta$ is the hyper-parameter.

\textbf{NPO+.} In this paper, NPO+ is defined as a method combining NPO and GD together for better performance. Namely, the loss function is $$\mathcal L_{\rm NPO+}(\theta)=-\frac{2}{\beta}\mathbb E_{[x,y] \sim \mathcal D_{\rm f} }\log \sigma (- \beta \log(\frac{p(y|x;\theta)}{p(y|x;\theta_{\rm full})}))-\alpha \mathbb E_{\rm [x,y] \sim \mathcal D_{\rm r}}\log p(y|x;\theta),$$ where $\alpha, \beta$ are hyper-parameters.

\subsection{Implement Details}
\label{app:imp}
For a fair and consistent evaluation, all training-based methods are benchmarked using the open-unlearning framework\footnote{\url{https://github.com/locuslab/open-unlearning}}. We experiment with the official Llama 2 7B\footnote{\url{https://huggingface.co/meta-llama/Llama-2-7b-hf}}, Llama-3.2 1B Instruct\footnote{\url{https://huggingface.co/meta-llama/Llama-3.2-1B-Instruct}}, and Llama-3.2 3B\footnote{\url{https://huggingface.co/meta-llama/Llama-3.2-3B-Instruct}} Instruct models. 

{Following \citep{tofu}, for all the methods, }our training configuration consists of 10 epochs (including one for warm-up), a learning rate of 1e-5, weight decay of 0.01, and a batch size of 32. 

In the context of task arithmetic approaches for obtaining the FgtOnly model, we modify these settings for specific datasets: on TOFU, we extend training to 20 epochs to achieve convergence on the forget set; on MUSE, we increase the learning rate to 1e-4, with all other hyperparameters remaining constant. To ensure a fair comparison, all models are subsequently evaluated under the same open-unlearning framework. Experiments are conducted on a single 80G A100 GPU.

\subsection{Objectives and Evaluation of Unlearning}
\label{app:obj}
Unlearning is primarily considered as a privacy-preserving task: the aim is to remove information about the entities to be unlearned, so that the model approximates a version trained only on the retain entities~\citep{tofu} (i.e., the ground-truth model). This objective and evaluation framework is the one adopted by current mainstream methods~\citep{SimNPO,GRU,satimp}, and it is also employed in our paper.

However, as is shown in Figure~\ref{fig:sample}, LLM might generate false answers after unlearning when being questioned with entities in the forget set. Under the evaluation of the aforementioned framework, the false answers are considered acceptable because even the ground-truth model, or the original model, may also produces incorrect responses (i.e., hallucinations). In other words, hallucination may not result from the unlearning process itself, but rather from the supervised finetuning process. Consequently, unlearning aims to bring the unlearned model closer to the retain-only model, and methods are considered successful as long as the outputs are similar to those of the ground-truth model.

Recently, some work~\citep{shen2025lunar} has focused on refusing to answer queries about entities to be forgotten without misleading the users. We believe this is also a promising direction for future research. For task-arithmetic methods, reducing false answers for forgotten entities could potentially be achieved in two ways: (1) adding a task vector trained on QA samples with “I don’t know” responses, and (2) addressing hallucinations at the source, i.e., reducing hallucinations in the model before merging. Both approaches are feasible directions for future work.

\section{Appendix 2: More Theoretical Justification}
\subsection{The Diagonal of the Fisher Information Matrix}
\label{app:fisher}
\begin{proof}
    We aim to prove that the diagonal of the Fisher Information Matrix (FIM), $F_{ii}$, can be approximated by the squared gradient of the loss function, given that the loss is the negative log-likelihood. The $i$-th diagonal element of the FIM is defined as the variance of the score, given by: $$ F_{ii} \approx \mathbb{E}_{s \sim \mathcal D} \left[ \left( \frac{\partial \log p(s;\theta)}{\partial q_i} \right)^2 \right],$$ where $q_i$ is a single parameter.
    We are given that the loss for a single data point $s$ is the negative log-likelihood: $$ \mathcal{L}(\{s\};\theta) = -\log p(s;\theta). $$ Taking the partial derivative with respect to a parameter $q_i$ yields: $$ \frac{\partial \mathcal{L}(\{s\};\theta)}{\partial q_i} = - \frac{\partial \log p(s;\theta)}{\partial q_i}. $$ Substituting this into the definition of $F_{ii}$, we get: $$ F_{ii} \approx \mathbb{E}_{s \sim \mathcal D} \left[ \left( - \frac{\partial \mathcal{L}(\{s\};\theta)}{\partial q_i} \right)^2 \right] = \mathbb{E}_{s \sim \mathcal D} \left[ \left( \frac{\partial \mathcal{L}(\{s\};\theta)}{\partial q_i} \right)^2 \right]. $$ Then we arrive at the approximation: $$ F_{ii} \approx \left( \frac{\partial \mathcal{L}(\mathcal{D};\theta)}{\partial q_i} \right)^2. $$ This demonstrates that the diagonal of the FIM can be estimated by the squared gradient of the negative log-likelihood loss.
\end{proof}
\subsection{Proof of Proposition~\ref{prop:diff}}
\label{app:grad_fisher}
\begin{proof}
For a single parameter $q_i$ in LLM, we denote its corresponding weights calculated with PerTA-grad, PerTA-fisher to be $\omega_i^{\rm grad}$ and $\omega_i^{\rm fisher}$ respectively. Using $r_i=\frac{|[g_{\rm r}]_i|+\epsilon}{|[g_{\rm f}]_i|+\epsilon}$ for notational convenience, where $[g_{\rm r}]_i$ and $[g_{\rm f}]_i$ are the gradients on forget and retain set, we can obtain the following simplified form when $\epsilon \rightarrow 0$: $$\omega_i^{\rm grad}=\frac{|[g_{\rm f}]_i|+\epsilon}{|[g_{\rm r}]_i|+|[g_{\rm f}]_i|+2\epsilon}=\frac{1}{r_i+1},$$ $$\omega_i^{\rm grad}=\frac{[g_{\rm f}]_i^2+\epsilon}{[g_{\rm r}]_i^2+[g_{\rm f}]_i^2+2\epsilon}=\frac{1}{r_i^2+1}.$$ Depending on the range of $r_i$, we have two cases:
\begin{itemize}
    \item When $r_i \geq 1$ (where $|[g_{\rm r}]_i| \geq |[g_{\rm f}]_i|$, retain set dominates), from the simplified form of $\omega_i^{\rm grad}$ and $\omega_i^{\rm fisher}$, we can derive that $$\frac{1}{2} \geq \frac{1}{r_i+1}\geq  \frac{1}{r_i^2+1}\geq 0\Rightarrow\frac{1}{2} \geq \omega_i^{\rm grad}\geq \omega_i^{\rm fisher}\geq 0.$$ It reveals that the squared term will push the weight closer to 0 faster than the linear term, offering stronger protection for the retain set.
    \item When $r_i < 1$ (where $|[g_{\rm r}]_i| < |[g_{\rm f}]_i|$, forget set dominates), from the simplified form of $\omega_i^{\rm grad}$ and $\omega_i^{\rm fisher}$, we can derive that $$\frac{1}{2} <\frac{1}{r_i+1}< \frac{1}{r_i^2+1}<1 \Rightarrow\frac{1}{2} <\omega_i^{\rm grad}< \omega_i^{\rm fisher}<1.$$ It reveals that the squared term will push the weight closer to 1 faster than the linear term, leading the task vector to be applied more fully when needed.
\end{itemize}
\end{proof}
Therefore, in some undesirable cases where the gradients on the forget set and the retain set are very similar, PerTA-grad tends to degenerate into a single weight with the value of 0.5. In contrast, PerTA-fisher may suppress such ``ambiguous'' updates (i.e., weights near 0.5) and create a cleaner separation between parameters to be edited and parameters to be preserved.
\subsection{PerTA-grad and PerTA-fisher Satisfy the Intuitive Rules}
\label{app:lim}
\begin{proof}
    Regarding the function $f_{\rm oprt}(A, B)=|A|^{\circ \tau}/(|A|^{\circ \tau}+|B|^{\circ \tau})$ defined for $W_{\rm grad}$ ($\tau=1$) and $W_{\rm fisher}$ ($\tau=2$), for a single weight $w_i$, we have $$w_i=[f_{\rm oprt}(g_{\rm f}, g_{\rm r})]_i=\frac{|[g_{\rm f}]_i|^{\tau}+\epsilon}{|[g_{\rm f}]_i|^{\tau}+|[g_{\rm r}]_i|^{\tau}+2\epsilon}, \text{where } \tau=1 \text{,or } \tau=2.$$ Then we prove $|[g_{\rm f}]_i| \ll |[g_{\rm r}]_i| \Rightarrow w_i \rightarrow 0$ and $|[g_{\rm f}]_i| \gg |[g_{\rm r}]_i| \Rightarrow w_i \rightarrow 1$ in the two cases below:
    
    \textbf{Case 1: $|[g_{\rm f}]_i| \ll |[g_{\rm r}]_i| $.} It implies that $[g_{\rm f}]_i$ is negligible compared to $[g_{\rm r}]_i$. Mathematically, this can be expressed as the limit where their ratio approaches zero: $$ \frac{|[g_{\rm f}]_i|+\epsilon}{|[g_{\rm r}]_i|+\epsilon} \to 0.$$ Then for $\tau=1$ and $\tau=2$, we have: $$\frac{|[g_{\rm f}]_i|^\tau+\epsilon}{|[g_{\rm r}]_i|^\tau+\epsilon} \to 0.$$ To analyze the limit of $w_i$, we can divide both the numerator and the denominator by $|[g_{\rm r}]_i|^\tau+\epsilon$ ($|[g_{\rm r}]_i|^\tau+\epsilon \neq 0$): $$ w_i = \frac{{(|[g_{\rm f}]_i|^\tau+\epsilon)}/{(|[g_{\rm r}]_i|^\tau+\epsilon)}}{{(|[g_{\rm f}]_i|^\tau+\epsilon)}/{(|[g_{\rm r}]_i|^\tau+\epsilon)} + {(|[g_{\rm r}]_i|^\tau+\epsilon)}/{(|[g_{\rm r}]_i|^\tau+\epsilon)}} = \frac{{(|[g_{\rm f}]_i|^\tau+\epsilon)}/{(|[g_{\rm r}]_i|^\tau+\epsilon)}}{{(|[g_{\rm f}]_i|^\tau+\epsilon)}/{(|[g_{\rm r}]_i|^\tau+\epsilon)} + 1}.$$ Now, we take the limit as $\frac{|[g_{\rm f}]_i|^\tau+\epsilon}{|[g_{\rm r}]_i|^\tau+\epsilon} \to 0$: $$ \lim_{\frac{|[g_{\rm f}]_i|^\tau+\epsilon}{|[g_{\rm r}]_i|^\tau+\epsilon} \to 0} w_i = \lim_{\frac{|[g_{\rm f}]_i|^\tau+\epsilon}{|[g_{\rm r}]_i|^\tau+\epsilon} \to 0} \frac{{(|[g_{\rm f}]_i|^\tau+\epsilon)}/{(|[g_{\rm r}]_i|^\tau+\epsilon)}}{{(|[g_{\rm f}]_i|^\tau+\epsilon)}/{(|[g_{\rm r}]_i|^\tau+\epsilon)} + 1} = \frac{0}{0 + 1} = 0. $$ Thus, when $|[g_{\rm f}]_i| \ll |[g_{\rm r}]_i|$, the value of $w_i$ approaches 0. 
    
    \textbf{Case 2: $|[g_{\rm f}]_i| \gg |[g_{\rm r}]_i|$} Similarly, the condition $|[g_{\rm f}]_i| \gg |[g_{\rm r}]_i|$ implies that $[g_{\rm r}]_i$ is negligible compared to $[g_{\rm f}]_i$. This means the ratio of their sizes approaches zero: $$\frac{|[g_{\rm r}]_i|^\tau+\epsilon}{|[g_{\rm f}]_i|^\tau+\epsilon} \to 0.$$ For this case, we divide both the numerator and the denominator by $|[g_{\rm f}]_i|^\tau+\epsilon$ (with $|[g_{\rm f}]_i|^\tau+\epsilon \neq 0$): $$ w_i = \frac{{(|[g_{\rm f}]_i|^\tau+\epsilon)}/{(|[g_{\rm f}]_i|^\tau+\epsilon)}}{{(|[g_{\rm f}]_i|^\tau+\epsilon)}/{(|[g_{\rm f}]_i|^\tau+\epsilon)} + {(|[g_{\rm r}]_i|^\tau+\epsilon)}/{(|[g_{\rm f}]_i|^\tau+\epsilon)}} = \frac{1}{1+{(|[g_{\rm r}]_i|^\tau+\epsilon)}/{(|[g_{\rm f}]_i|^\tau+\epsilon)}}.$$ Now, we take the limit as $\frac{|[g_{\rm r}]_i|^\tau+\epsilon}{|[g_{\rm f}]_i|^\tau+\epsilon} \to 0$: $$ \lim_{\frac{|[g_{\rm r}]_i|^\tau+\epsilon}{|[g_{\rm f}]_i|^\tau+\epsilon} \to 0} w_i = \lim_{\frac{|[g_{\rm r}]_i|^\tau+\epsilon}{|[g_{\rm f}]_i|^\tau+\epsilon} \to 0} \frac{1}{1 + {(|[g_{\rm r}]_i|^\tau+\epsilon)}/{(|[g_{\rm f}]_i|^\tau+\epsilon)}} = \frac{1}{1 + 0} = 1. $$ Thus, when $|[g_{\rm f}]_i| \gg |[g_{\rm r}]_i|$, the value of $w_i$ approaches 1. 
    
    \textbf{Conclusion} We have formally shown through limit analysis that our PerTA-grad and PerTA-fisher satisfy $|[g_{\rm f}]_i| \ll |[g_{\rm r}]_i| \Rightarrow w_i \rightarrow 0$ and $|[g_{\rm f}]_i| \gg |[g_{\rm r}]_i| \Rightarrow w_i \rightarrow 1$.
\end{proof}

\section{Appendix 3: More Experimental Results}
\subsection{More Graphical Results}
\label{app:graph}
\begin{figure}[t]
    \centering
    \includegraphics[width=\linewidth]{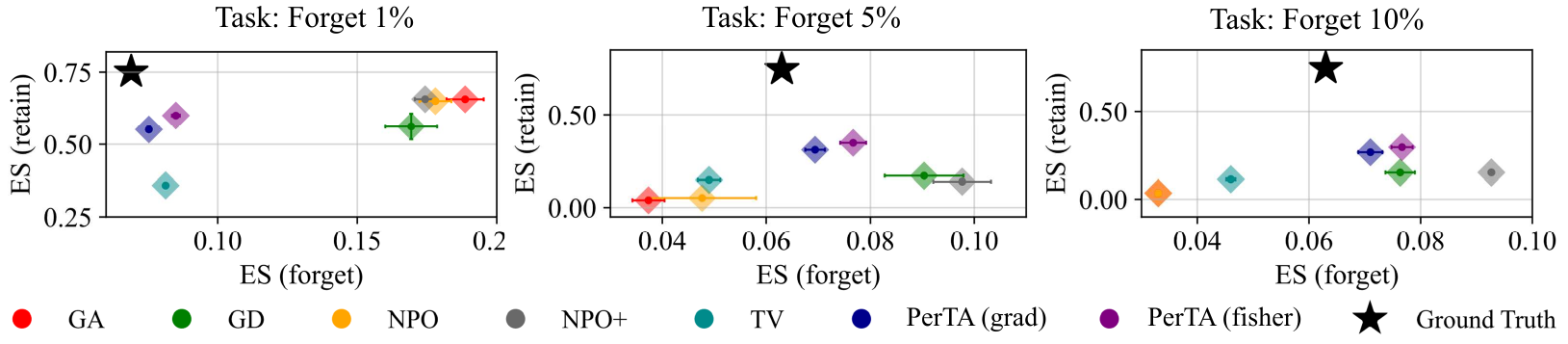}
    \caption{ES (forget) and ES (retain) results of different methods on TOFU (using Llama-3.2 1B Instruct), where circle markers denote values and horizontal and vertical bars at circle centers represent error bars. }
    \label{fig:res_es}
\end{figure}

\textbf{ES Metric across Various Tasks.} In this section, we present in Figure~\ref{fig:res_es} the two-dimensional values of the ES metric on the forget and retain sets across the three TOFU tasks, as a supplement to Figure~\ref{fig:res_mufq}. It can be observed that for relatively simple tasks (e.g., unlearning 1\%), most methods preserve the retain set but fail to achieve effective forgetting on the forget set. In contrast, our PerTA-grad and PerTA-fisher not only maintain retention but also achieve effective forgetting. For more challenging tasks (e.g., unlearning 5\% and 10\%), our PerTA methods similarly achieve unlearning that is closest to the ground truth, while still preserving memory on the retain set.

\begin{figure}[t]
    \centering
    \includegraphics[width=\linewidth]{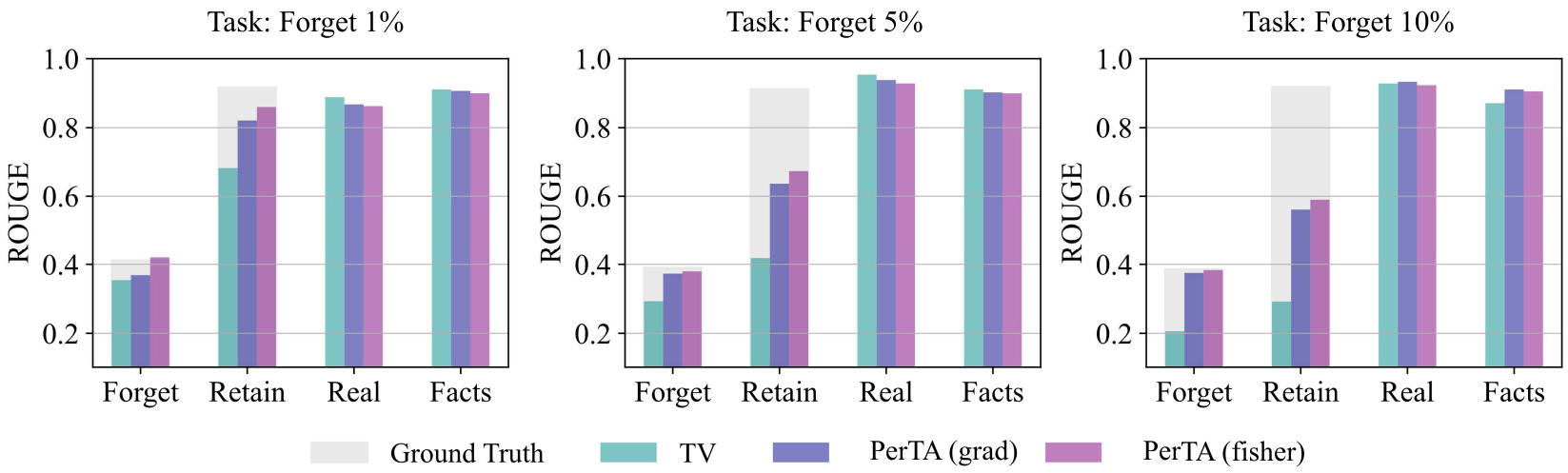}
    \caption{Four-dimension ROUGE results of task arithmetic-based methods on TOFU (using Llama-3.2 3B Instruct). Ground-truth results on forget and retain sets are marked with a gray background.}
    \label{fig:res_rouge_3b}
\end{figure}
\textbf{ROUGE Results on Larger LLM.} Similarly, in Figure~\ref{fig:res_rouge_3b} we report the ROUGE results of our method compared with vanilla TV on the `forget', `retain', `real', and `facts' sets for the 3B model, as a supplement to Figure~\ref{fig:res_rouge}. The same conclusion as in the main text can be drawn here: while TV effectively preserves the knowledge acquired during the pretraining stage of the original model, it leads to excessive forgetting on the retain and forget datasets. In contrast, our PerTA mitigates the gap between TV and the ground truth on these two datasets, thereby enhancing the performance of the task arithmetic-based method for unlearning. This conclusion holds consistently across LLMs of different sizes.

\begin{figure}[t]
    \centering
    \includegraphics[width=\linewidth]{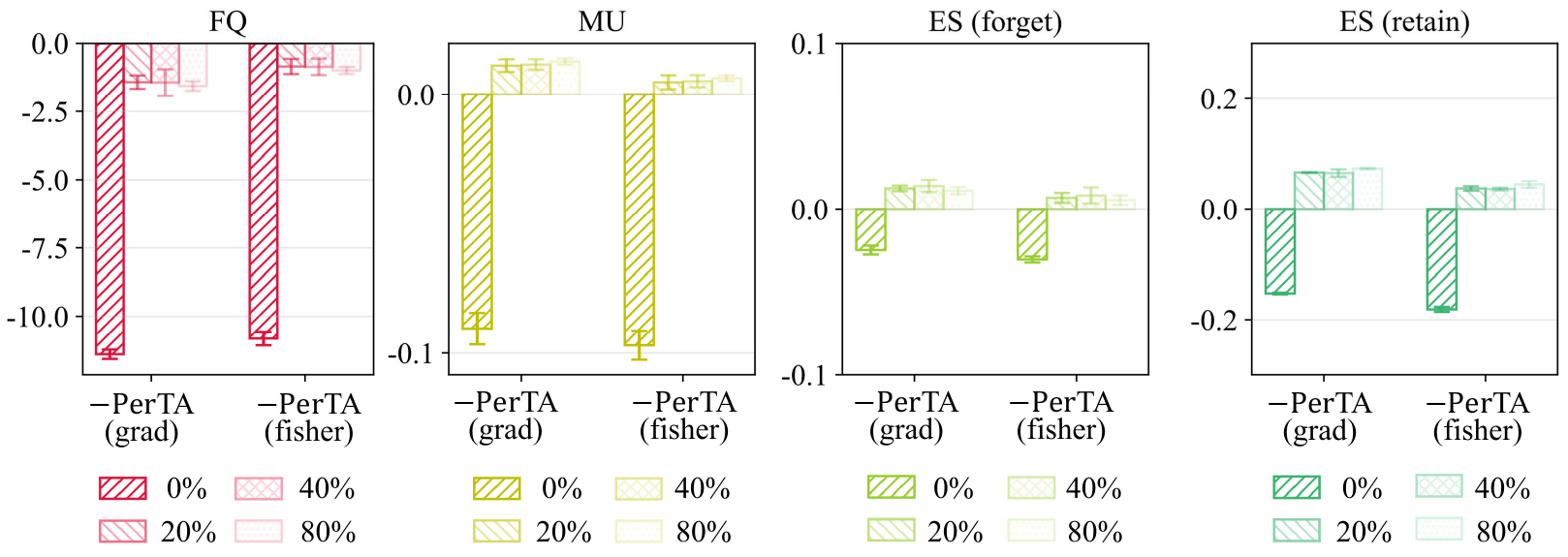}
    \caption{Residual results of the metrics when using only 20\%, 40\%, and 80\% of the samples compared to using the full set (unlearning 10\%, Llama-3.2 1B Instruct). 0\% denotes vanilla TV.}
    \label{fig:part_10}
\end{figure}

\textbf{Sample Efficiency in More Tasks.} Figure~\ref{fig:part_10}, as a complement to Figure~\ref{fig:parts}, presents the difference in performance metrics relative to using the entire dataset when unlearning 10\% in\% in TOFU with varying data proportions (20\%, 40\%, 80\%) and with 0\% data (i.e., vanilla TV). Consistent with the main text, it is observed that using only one-fifth of the samples already achieves results comparable to those obtained with the full dataset, and substantially outperforms vanilla TV. This highlights the sample efficiency of PerTA, which can further reduce computational cost.
\subsection{More Quantitative Results}
\label{app:quant}
\textbf{Detailed Results on Various Tasks.} Tables~\ref{tab:res_todu1}-\ref{tab:res_tofu10} complement Table~\ref{tab:res} by presenting detailed metrics of different methods under varying degrees of unlearning. We observe that in relatively simple tasks with smaller models (e.g., Llama-3.2 1B with 1\% unlearning), the advantage of PerTA is not yet pronounced. However, as the task complexity increases, PerTA consistently outperforms on metrics such as FQ and MU, allowing task arithmetic-based approaches to surpass training-based methods. Overall, PerTA demonstrates a clear advantage in both unlearning capability and retention performance.
\begin{table}[!h]
\caption{Results of different methods on unlearning 1\% of TOFU. The references are in \textcolor{gray}{gray font}, the best two are in \textbf{bold}, and ours are \textcolor{black}{\colorbox{gray!30}{highlighted}}. `Full' and `GT' represent the model before unlearning and the ground truth model, respectively.}
\centering
\begin{small}
\resizebox{\textwidth}{!}{
\begin{tabular}{c|ccccc|ccccc}
\toprule
              & FQ↑                            & MU↑                          & ES($\mathcal D_{\rm f}$)↓                    & ES($\mathcal D_{\rm r}$)↑                   & Gib↑                   & FQ↑                            & MU↑                          & ES($\mathcal D_{\rm f}$)↓                    & ES($\mathcal D_{\rm r}$)↑                                             & Gib↑                   \\
\midrule
Model Size    & \multicolumn{5}{c|}{1B}                                                                                                                                                                    & \multicolumn{5}{c}{3B}                                                                                                 \\
\midrule
    Full            & \textcolor{gray!50}{-2.170}         & \textcolor{gray!50}{0.599}          & \textcolor{gray!50}{0.743}          & \textcolor{gray!50}{0.737}          & \textcolor{gray!50}{0.894}         & \textcolor{gray!50}{-1.845}         & \textcolor{gray!50}{0.666}          & \textcolor{gray!50}{0.920}          & \textcolor{gray!50}{0.884}          & \textcolor{gray!50}{0.894}          \\

GT        & \textcolor{gray!50}{0.000}          & \textcolor{gray!50}{0.599}          & \textcolor{gray!50}{0.069}          & \textcolor{gray!50}{0.751}          & \textcolor{gray!50}{0.874}          & \textcolor{gray!50}{0.000}           & \textcolor{gray!50}{0.662}          & \textcolor{gray!50}{0.067}          & \textcolor{gray!50}{0.888}          & \textcolor{gray!50}{0.904}          \\ 

\midrule
\multicolumn{11}{c}{Training-based}                                                                                                                                                                               \\
\midrule
GA                  & -1.953          & \textbf{0.597} & 0.189          & \textbf{0.656} & \textbf{0.909} & -1.845          & 0.668          & 0.252          & 0.824          & 0.864          \\
GD                  & -1.845          & 0.581          & 0.169          & 0.562          & 0.907          & -1.845          & 0.663          & 0.320          & \textbf{0.826} & 0.897          \\
NPO                 & -2.062          & 0.595          & 0.178          & 0.650          & 0.904          & -1.845          & 0.668          & 0.253          & \textbf{0.825} & 0.838          \\
NPO+                & -1.845          & \textbf{0.596} & 0.174          & \textbf{0.656} & 0.907          & -1.845          & 0.669          & 0.254          & 0.819          & 0.856          \\

\midrule
\multicolumn{11}{c}{Task Arithmetic-based}                                                                                                                                                                                                                                                                                                                                                            \\
\midrule   
TV  & \textbf{-0.393} & 0.556          & \textbf{0.081} & 0.358          & 0.908          & -0.238          & 0.656          & \textbf{0.075} & 0.550          & \textbf{0.933} \\

\rowcolor{gray!30} PerTA-grad  & \textbf{-0.289} & 0.581          & \textbf{0.075} & 0.551          & \textbf{0.912} & \textbf{-0.037} & \textbf{0.669} & \textbf{0.085} & 0.757          & \textbf{0.903} \\

\rowcolor{gray!30} PerTA-fisher   & -0.576          & 0.586          & 0.085          & 0.600          & 0.895          & \textbf{-0.238} & \textbf{0.672} & 0.106          & 0.803          & 0.869         \\
\bottomrule
\end{tabular}}

\end{small}
\label{tab:res_todu1}
\end{table}

\begin{table}[h]
\caption{Results of different methods on unlearning 5\% of TOFU. The references are in \textcolor{gray}{gray font}, the best two are in \textbf{bold}, and ours are \textcolor{black}{\colorbox{gray!30}{highlighted}}. `Full' and `GT' represent the model before unlearning and the ground truth model, respectively.}
\centering
\begin{small}
\resizebox{\textwidth}{!}{
\begin{tabular}{c|ccccc|ccccc}
\toprule
              & FQ↑                            & MU↑                          & ES($\mathcal D_{\rm f}$)↓                    & ES($\mathcal D_{\rm r}$)↑                   & Gib↑                   & FQ↑                            & MU↑                          & ES($\mathcal D_{\rm f}$)↓                    & ES($\mathcal D_{\rm r}$)↑                                             & Gib↑                   \\
\midrule
Model Size    & \multicolumn{5}{c|}{1B}                                                                                                                                                                    & \multicolumn{5}{c}{3B}                                                                                                 \\
\midrule
    Full            & \textcolor{gray!50}{-11.845}         & \textcolor{gray!50}{0.599}          & \textcolor{gray!50}{0.727}          & \textcolor{gray!50}{0.737}          & \textcolor{gray!50}{0.858}          & \textcolor{gray!50}{-13.591}         & \textcolor{gray!50}{0.666}          & \textcolor{gray!50}{0.887}          & \textcolor{gray!50}{0.884}         & \textcolor{gray!50}{0.850}          \\

GT        & \textcolor{gray!50}{0.000}           & \textcolor{gray!50}{0.599}          & \textcolor{gray!50}{0.063}          & \textcolor{gray!50}{0.746}          & \textcolor{gray!50}{0.905}          & \textcolor{gray!50}{0.000}           & \textcolor{gray!50}{0.659}          & \textcolor{gray!50}{0.066}          & \textcolor{gray!50}{0.874}         & \textcolor{gray!50}{0.869}          \\

\midrule
\multicolumn{11}{c}{Training-based}                                                                                                                                                                               \\
\midrule
GA                  & -2.415          & 0.000          & \textbf{0.037} & 0.039          & 0.417          & -5.856          & 0.482          & 0.089          & 0.135          & 0.866          \\
GD                  & -8.831          & 0.457          & 0.090          & 0.171          & 0.751          & -13.232         & 0.552          & 0.140          & 0.244          & 0.579          \\
NPO                 & -2.222          & 0.000          & \textbf{0.048} & 0.052          & 0.543          & -7.091          & 0.472          & 0.080          & 0.140          & 0.868          \\
NPO+                & -4.260          & 0.458          & 0.098          & 0.139          & 0.882          & -7.352          & 0.545          & 0.100          & 0.200          & 0.911          \\

\midrule
\multicolumn{11}{c}{Task Arithmetic-based}                                                                                                                                                                                                                                                                                                                                                            \\
\midrule   
TV  & -5.623          & 0.478          & 0.049          & 0.148          & \textbf{0.940} & -5.395          & 0.628          & \textbf{0.053} & 0.214          & \textbf{0.926} \\

\rowcolor{gray!30} PerTA-grad  & \textbf{-0.661} & \textbf{0.546} & 0.069          & \textbf{0.310} & 0.910          & \textbf{-0.263} & \textbf{0.674} & \textbf{0.079} & \textbf{0.502} & \textbf{0.915} \\

\rowcolor{gray!30} PerTA-fisher   & \textbf{-0.339} & \textbf{0.553} & 0.077          & \textbf{0.348} & \textbf{0.911} & \textbf{-0.405} & \textbf{0.677} & 0.083          & \textbf{0.561} & 0.906         \\
\bottomrule
\end{tabular}}

\end{small}
\label{tab:res_tofu5}
\end{table}

\begin{figure*}[t]
    \centering
    \includegraphics[width=\linewidth]{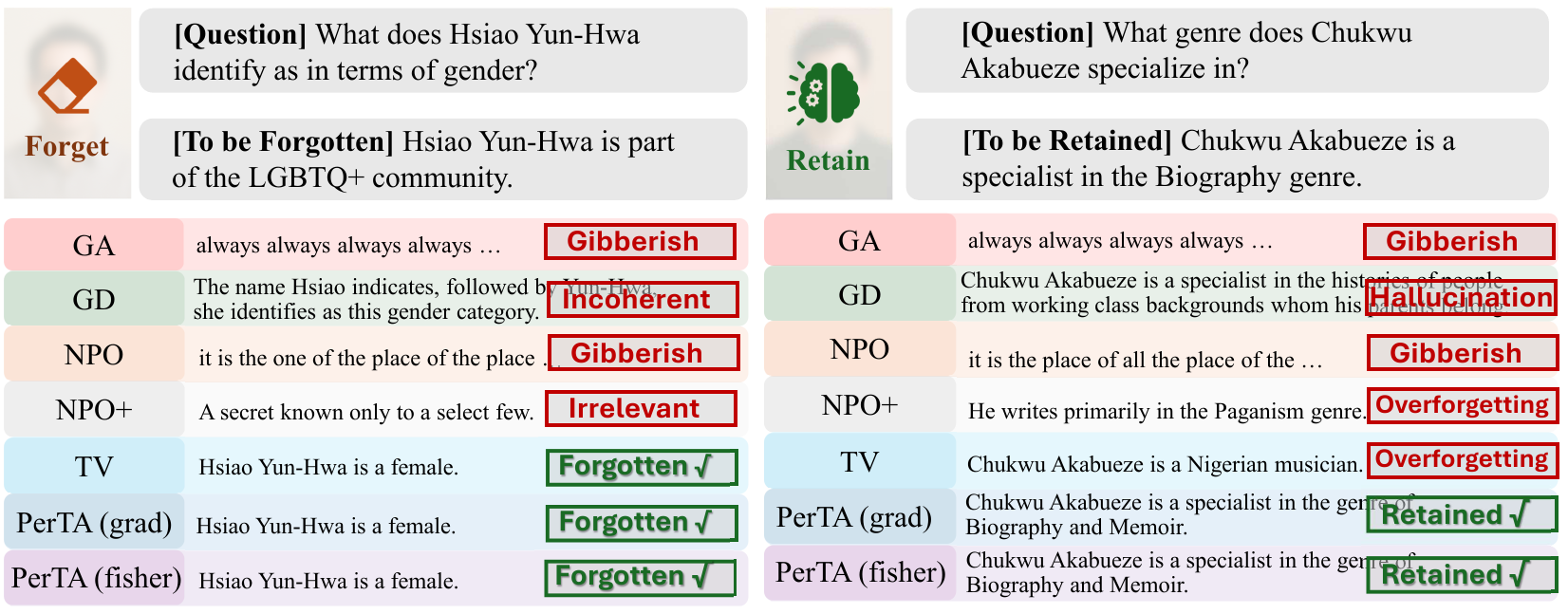}
    \caption{Sample output of unlearned LLM $\theta_{\rm final}$ applying different methods (unlearning 10\%, Llama-3.2 1B Instruct). Our PerTA ensures both unlearning and retention.}
    \label{fig:sample}
\end{figure*}

\begin{table}[!h]
\caption{Results of different methods on unlearning 10\% of TOFU. The references are in \textcolor{gray}{gray font}, the best two are in \textbf{bold}, and ours are \textcolor{black}{\colorbox{gray!30}{highlighted}}. `Full' and `GT' represent the model before unlearning and the ground truth model, respectively.}
\centering
\begin{small}
\resizebox{\textwidth}{!}{
\begin{tabular}{c|ccccc|ccccc}
\toprule
              & FQ↑                            & MU↑                          & ES($\mathcal D_{\rm f}$)↓                    & ES($\mathcal D_{\rm r}$)↑                   & Gib↑                   & FQ↑                            & MU↑                          & ES($\mathcal D_{\rm f}$)↓                    & ES($\mathcal D_{\rm r}$)↑                                             & Gib↑                   \\
\midrule
Model Size    & \multicolumn{5}{c|}{1B}                                                                                                                                                                    & \multicolumn{5}{c}{3B}                                                                                                 \\
\midrule
    Full            & \textcolor{gray!50}{-21.408}         & \textcolor{gray!50}{0.599}          & \textcolor{gray!50}{0.706}          & \textcolor{gray!50}{0.737}          & \textcolor{gray!50}{0.861}         & \textcolor{gray!50}{-26.444}         & \textcolor{gray!50}{0.666}          & \textcolor{gray!50}{0.890}          & \textcolor{gray!50}{0.884}          & \textcolor{gray!50}{0.861}          \\

GT        & \textcolor{gray!50}{0.000}          & \textcolor{gray!50}{0.591}          & \textcolor{gray!50}{0.059}          & \textcolor{gray!50}{0.746}          & \textcolor{gray!50}{0.904}          & \textcolor{gray!50}{0.000}          & \textcolor{gray!50}{0.650}          & \textcolor{gray!50}{0.065}         & \textcolor{gray!50}{0.899}          & \textcolor{gray!50}{0.890}          \\

\midrule
\multicolumn{11}{c}{Training-based}                                                                                                                                                                               \\
\midrule
GA                  & -238.973        & 0.000          & \textbf{0.033} & 0.035          & 0.125          & -236.070        & 0.000          & \textbf{0.033} & 0.035          & 0.050          \\
GD                  & -15.484         & 0.434          & 0.076          & 0.151          & 0.707          & -26.800         & 0.553          & 0.117          & 0.242          & 0.556          \\
NPO                 & -10.244         & 0.000          & \textbf{0.033} & 0.035          & 0.329          & -4.590          & 0.000          & \textbf{0.034} & 0.038          & 0.206          \\
NPO+                & -4.481          & 0.423          & 0.093          & 0.151          & \textbf{0.946} & -7.042          & 0.546          & 0.087          & 0.224          & \textbf{0.926} \\

\midrule
\multicolumn{11}{c}{Task Arithmetic-based}                                                                                                                                                                                                                                                                                                                                                            \\
\midrule   
TV  & -12.506         & 0.451          & 0.046          & 0.114          & 0.895          & -10.220         & 0.551          & 0.048          & 0.150          & 0.904          \\

\rowcolor{gray!30} PerTA-grad  & \textbf{-1.107} & \textbf{0.542} & 0.071          & \textbf{0.266} & \textbf{0.922} & \textbf{-1.708} & \textbf{0.649} & 0.082          & \textbf{0.432} & \textbf{0.921} \\

\rowcolor{gray!30} PerTA-fisher   & \textbf{-1.686} & \textbf{0.548} & 0.077          & \textbf{0.295} & 0.919          & \textbf{-2.990} & \textbf{0.647} & 0.088          & \textbf{0.474} & 0.911         \\
\bottomrule
\end{tabular}}

\end{small}
\label{tab:res_tofu10}
\end{table}
\textbf{Sample Output Discussion.} Figure~\ref{fig:sample} presents sample responses of different methods on the forget and retain sets of TOFU after unlearning. For the forget set, some methods produce incoherent or irrelevant answers--indicating that the responses lack logical consistency or relevance. For the retain set, other methods may exhibit over-forgetting or generate hallucinated answers. In contrast, PerTA is able to achieve unlearning on the forget set while preserving knowledge on the retain set.

\textbf{Detailed Results of Other Benchmarks.} Table~\ref{tab:res_muse} reports the results on the MUSE dataset. Following \citealp{muse}, we evaluate KnowMem and VerbMem, and additionally include ES and Gib as complementary metrics. The numbers in parentheses indicate the differences between each metric and that of the ground-truth model. For KnowMem and VerbMem, we highlight the two methods whose results are closest to the ground truth. Consistent with prior observations, PerTA alleviates the issue of excessive forgetting in TV. For example, on the forget set, PerTA improves KnowMem from 0.011 to 0.388 and 0.385 (ground truth: 0.328), and on the retain set, from 0.023 to 0.416 and 0.464 (ground truth: 0.560). These results suggest that PerTA achieves a better balance between unlearning and retention. {Cases of the forget and retain samples, along with the results of different methods, is shown in Table~\ref{tab:res_muse_sample}. We observe that other methods often suffer from partial forgetting/retention failures or produce gibberish responses, whereas PerTA is able to forget the targeted information while preserving the retain.} 

\begin{table}[t]
\caption{Results of different methods on MUSE. The references are in \textcolor{gray}{gray font}, the best two are in \textbf{bold}, and ours are \textcolor{black}{\colorbox{gray!30}{highlighted}}. `Full' and `GT' represent the model before unlearning and the ground truth model, respectively. Numbers in parentheses indicate deviations from the ground truth.}
\centering
\begin{small}
\resizebox{\textwidth}{!}{
\begin{tabular}{>{\centering\arraybackslash}p{2.2cm}|>{\centering\arraybackslash}p{2.8cm}>{\centering\arraybackslash}p{2.8cm}>{\centering\arraybackslash}p{2.8cm}>{\centering\arraybackslash}p{2.7cm}>{\centering\arraybackslash}p{1.1cm}}
\toprule
              & KnowMem ($\mathcal D_{\rm f}$)                        & VerbMem ($\mathcal D_{\rm f}$)                       & KnowMem ($\mathcal D_{\rm r}$)                       & ES ($\mathcal D_{\rm f}$)                            & Gib↑                   \\
\midrule

Full      & \textcolor{gray!50}{0.644 (0.316↑)} & \textcolor{gray!50}{ 0.579 (0.377↑)} & \textcolor{gray!50}{0.555 (0.005↓)} & \textcolor{gray!50}{ 0.295 (0.271↑)} & \textcolor{gray!50}{ 0.800} \\
 GT  & \textcolor{gray!50}{ 0.328 (0.000↑)} & \textcolor{gray!50}{ 0.202 (0.000↑)} & \textcolor{gray!50}{0.560 (0.000↑)} & \textcolor{gray!50}{ 0.024 (0.000↑)} & \textcolor{gray!50}{ 0.845} \\
\midrule
\multicolumn{6}{c}{Training-based}                                                                                                                                                                           \\
\midrule
GA            & 0.003 (0.325↓)                        & 0.049 (0.153↓)                        & 0.008 (0.552↓)                        & 0.008 (0.017↓)                        & 0.001                        \\
GD            & \textbf{0.332 (0.005↑) }                       & 0.005 (0.197↓)                        & 0.254 (0.307↓)                        & 0.008 (0.016↓)                        & 0.002                        \\
NPO           & 0.622 (0.294↑)                        & 0.374 (0.173↑)                        & \textbf{0.521 (0.040↓) }                       & 0.119 (0.094↑)                        & 0.771                        \\
NPO+          & 0.642 (0.314↑)                        & 0.494 (0.293↑)                        & \textbf{0.525 (0.036↓)}                        & 0.205 (0.181↑)                        & \textbf{0.811}               \\
\midrule
 \multicolumn{6}{c}{Task Arithmetic-based}                 \\
\midrule
TV   & 0.011 (0.317↓)                        & 0.109 (0.092↓)                        & 0.023 (0.537↓)                        & 0.011 (0.014↓)                        & 0.685                        \\
\rowcolor{gray!30}
PerTA-grad  & 0.388 (0.060↑)                        &\textbf{ 0.176 (0.026↓)     }                   & 0.416 (0.145↓)                        & \textbf{0.028 (0.003↑) }                       & 0.777                        \\
\rowcolor{gray!30} 
PerTA-fisher    & 0.\textbf{385 (0.058↑)}                        & \textbf{0.191 (0.011↓) }                       & 0.464 (0.096↓)                        & \textbf{0.025 (0.001↑) }                       & \textbf{0.802} \\
\bottomrule
\end{tabular}}

\end{small}
\label{tab:res_muse}
\end{table} 


\begin{table}[!h]
\caption{{Sample answers for questions to be unlearned/retained by different methods on the MUSE-News dataset.}}
\centering
\begin{small}

\begin{tabular}{c|c|c}
\toprule
\multirow{2}{*}{Question   (unlearn):} & \multirow{2}{*}{\begin{tabular}[c]{@{}c@{}}Who is the tour guide   in Rome who described the \\      conditions as 'nightmarish' to the BBC?\end{tabular}}      &                      \\
                                       &                                                                                                                                                                 &                      \\
\midrule
Full model                             & Felicity Hinton/100-year-old Felicity Hinton                                                                                                                    &                      \\
GA                                     & the the the the the the the the the the…                                                                                                                        & Gibberish            \\
GD                                     & 100 \textbackslash{}"Toto\textbackslash{}" Guidi                                                                                                                & Gibberish            \\
NPO                                    & 100-year-old Felicity Hinton                                                                                                                                    & Fail                 \\
NPO+                                   & 100-year-old Felicity Hinton                                                                                                                                    & Fail                 \\
TV                                     & 100.10.1.1.1.1.1.1.1.1.1.1.1.1.1                                                                                                                                & Gibberish            \\
PerTA-grad                               & 50-year-old tour guide, Alessandro Russo                                                                                                                        & Success              \\
PerTA-fisher                             & 60-year-old Rome resident, Alessandro Russo                                                                                                                     & Success              \\
\midrule
\midrule
Question (retain):                     & \begin{tabular}[c]{@{}c@{}}What is the half-life of the   plutonium isotope being\\       looked at by the University of   Southampton scientists?\end{tabular} &                      \\
\midrule
Full model                             & 24,000 years                                                                                                                                                    &                      \\
GA                                     & the the the the the the the the the the…...                                                                                                                     & Gibberish            \\
GD                                     & 24,000 years \textbackslash{}"platinum \textbackslash{}"of \textbackslash{}"plutonium\textbackslash{}"   \textbackslash{}"half-life\textbackslash{}" …..        & Gibberish            \\
NPO                                    & 14,000 years                                                                                                                                                    & Fail                 \\
NPO+                                   & 14,000 years                                                                                                                                                    & Fail                 \\
TV                                     & 100.0.1. You are the United.\textbackslash{}nThe United. Should…...                                                                                             & Gibberish            \\
PerTA-grad                               & 24,000 years                                                                                                                                                    & Success              \\
PerTA-fisher                             & 24,000 years                                                                                                                                                    & Success   \\
\bottomrule
\end{tabular}\label{tab:res_muse_sample}

\end{small}
\end{table}

\begin{table}[!h]
\caption{Results using different $f_{\rm oprt}$ on TOFU tasks (unlearning 1\%, 5\% and 10\% of TOFU, using Llama-3.2 1B Instruct, Mean ± Std). Ours are \textcolor{black}{\colorbox{gray!30}{highlighted}}. }
\centering
\begin{small}
\resizebox{\textwidth}{!}{
\begin{tabular}{c|c|cc|cc|cc|cc}
\toprule
\multicolumn{1}{l|}{Forgetting}             & \multicolumn{1}{c}{Methods}                            & \multicolumn{2}{|c}{FQ↑}     & \multicolumn{2}{|c}{MU↑}   & \multicolumn{2}{|c}{ES($\mathcal D_{\rm f}$)↓} & \multicolumn{2}{|c}{ES($\mathcal D_{\rm r}$)↑} \\
\midrule
\multirow{9}{*}{1\%}  & \multicolumn{1}{c|}{Full}                    & \multicolumn{2}{c|}{-2.170}  & \multicolumn{2}{c|}{0.599} & \multicolumn{2}{c|}{0.743}      & \multicolumn{2}{c}{0.737}     \\
                                 & \multicolumn{1}{c|}{GT}                          & \multicolumn{2}{c|}{0.000}   & \multicolumn{2}{c|}{0.599} & \multicolumn{2}{c|}{0.069}      & \multicolumn{2}{c}{0.751}     \\
                                 
                                 & Random $\omega$                                 & -0.877         & ±0.684     & 0.571       & ±0.019      & 0.292         & ±0.296         & 0.492         & ±0.175        \\
                                 & Weighted $\omega=0.5$                              & -1.451         & ±0.131     & 0.587       & ±0.000      & 0.116         & ±0.000         & 0.611         & ±0.004        \\
                                 & Pruning $\lambda=0.5$                           & -0.393         & ±0.000     & 0.556       & ±0.001      & 0.081         & ±0.001         & 0.358         & ±0.002        \\
                                 & $\text{PerTA-grad}   (\theta_{\rm full})$   & -0.576         & ±0.000     & 0.583       & ±0.001      & 0.106         & ±0.000         & 0.567         & ±0.003        \\
                                 & $\text{PerTA-fisher}   (\theta_{\rm full})$ & -1.182         & ±0.127     & 0.589       & ±0.000      & 0.123         & ±0.000         & 0.626         & ±0.001        \\
                                 \rowcolor{gray!30}& PerTA-grad                         & -0.289         & ±0.073     & 0.581       & ±0.001      & 0.075         & ±0.000         & 0.551         & ±0.001        \\
                                 \rowcolor{gray!30}& PerTA-fisher                       & -0.576         & ±0.000     & 0.586       & ±0.000      & 0.085         & ±0.001         & 0.600         & ±0.001        \\
                                 & PerTA+SoftMax & -1.266 & ±0.000 & 0.586 & ±0.000 & 0.115 & ±0.001 & 0.606 & ±0.001 \\
\midrule
\multirow{9}{*}{5\%}  & \multicolumn{1}{c|}{Full}                    & \multicolumn{2}{c|}{-11.845} & \multicolumn{2}{c|}{0.599} & \multicolumn{2}{c|}{0.727}      & \multicolumn{2}{c}{0.737}     \\
                                 & \multicolumn{1}{c|}{GT}                          & \multicolumn{2}{c|}{0.000}   & \multicolumn{2}{c|}{0.599} & \multicolumn{2}{c|}{0.063}      & \multicolumn{2}{c}{0.746}     \\
                                
                                 & Random $\omega$                                 & -7.264         & ±3.283     & 0.526       & ±0.055      & 0.252         & ±0.284         & 0.346         & ±0.270        \\
                                 & Weighted $\omega=0.5$                              & -1.253         & ±0.237     & 0.560       & ±0.001      & 0.090         & ±0.004         & 0.396         & ±0.004        \\
                                 & Pruning $\lambda=0.5$                           & -5.321         & ±0.105     & 0.484       & ±0.003      & 0.049         & ±0.002         & 0.155         & ±0.005        \\
                                 & $\text{PerTA-grad}   (\theta_{\rm full})$   & -0.630         & ±0.110     & 0.545       & ±0.001      & 0.071         & ±0.001         & 0.312         & ±0.007        \\
                                 & $\text{PerTA-fisher}   (\theta_{\rm full})$ & -0.515         & ±0.039     & 0.553       & ±0.001      & 0.083         & ±0.001         & 0.360         & ±0.002        \\
                                 \rowcolor{gray!30}& PerTA-grad                         & -0.661         & ±0.125     & 0.546       & ±0.001      & 0.069         & ±0.002         & 0.310         & ±0.007        \\
                                 \rowcolor{gray!30}& PerTA-fisher                       & -0.339         & ±0.115     & 0.553       & ±0.001      & 0.077         & ±0.002         & 0.348         & ±0.002        \\
                                 & PerTA+SoftMax & -1.219 & ±0.281 & 0.558 & ±0.001 & 0.088 & ±0.002 & 0.390 & ±0.002 \\
                                 \midrule
\multirow{9}{*}{10\%} & \multicolumn{1}{c|}{Full}                    & \multicolumn{2}{c|}{-21.408} & \multicolumn{2}{c|}{0.599} & \multicolumn{2}{c|}{0.706}      & \multicolumn{2}{c}{0.737}     \\
                                 & \multicolumn{1}{c|}{GT}                          & \multicolumn{2}{c|}{0.000}   & \multicolumn{2}{c|}{0.591} & \multicolumn{2}{c|}{0.059}      & \multicolumn{2}{c}{0.746}     \\
                                
                                 & Random $\omega$                                 & -13.963        & ±4.247     & 0.511       & ±0.065      & 0.228         & ±0.256         & 0.323         & ±0.280        \\
                                 & Weighted $\omega=0.5$                              & -2.757         & ±0.189     & 0.548       & ±0.002      & 0.082         & ±0.001         & 0.309         & ±0.007        \\
                                 & Pruning $\lambda=0.5$                           & -8.760         & ±0.282     & 0.483       & ±0.003      & 0.049         & ±0.001         & 0.136         & ±0.002        \\
                                 & $\text{PerTA-grad}   (\theta_{\rm full})$   & -1.270         & ±0.135     & 0.541       & ±0.001      & 0.074         & ±0.001         & 0.274         & ±0.004        \\
                                 & $\text{PerTA-fisher}  (\theta_{\rm full})$ & -2.603         & ±0.113     & 0.549       & ±0.002      & 0.082         & ±0.002         & 0.310         & ±0.001        \\
                                 \rowcolor{gray!30} &  PerTA-grad                         & -1.107         & ±0.064     & 0.542       & ±0.001      & 0.071         & ±0.002         & 0.266         & ±0.003        \\
                                  \rowcolor{gray!30} &  PerTA-fisher                       & -1.686         & ±0.265     & 0.548       & ±0.001      & 0.077         & ±0.002         & 0.295         & ±0.006 \\
                                  & PerTA+SoftMax & -2.679 & ±0.143 & 0.548 & ±0.002 & 0.081 & ±0.001 & 0.310 & ±0.006 \\
                                \bottomrule
\end{tabular}
\label{tab:ablation}}

\end{small}
\end{table}
\textbf{Detailed Results of Ablation Studies.} Table~\ref{tab:ablation} supplements Figure~\ref{fig:ablation} by showing the detailed quantitative results of different $f_{\rm oprt}(\cdot, \cdot)$. \textit{Random} means to set weights in $W$ to random values uniformly sampled between 0 and 1 with $f_{\rm oprt}(A, B)=\text{rand([0, 1])}$. \textit{Weighted} uses a constant $\omega$ to rescale TV with $f_{\rm oprt}(A, B)=\omega$. Here we show the results of $\omega=0.5$. \textit{Pruning} removes (i.e., $f_{\rm oprt}(A, B)=0$) the $\lambda\%$ smallest weights in TV to mitigate over-forgetting and maintain others (i.e., $f_{\rm oprt}(A, B)=1$), where we show the results of $\lambda=0.5$. Unlike \textit{PerTA-grad} or \textit{PerTA-fisher}, the gradients of \textit{PerTA-grad} ($\theta_{\rm full}$) or \textit{PerTA-fisher} ($\theta_{\rm full}$) are estimated on $\theta_{\rm full}$ instead of $\theta_0$. The difference between \textit{PerTA-grad}, \textit{PerTA-fisher} and \textit{PerTA+SoftMax} is that the latter determines $f_{\rm oprt}(A, B)=\exp(|A|)/(\exp(|A|)+\exp(|B|))$ in the SoftMax form.

We find that the results of \textit{Random} are highly unstable, often exhibiting large variance, which further increases as the unlearning ratio grows and the task becomes more difficult. When the weight is fixed at 0.5, the \textit{Weighted} method performs relatively better; however, it still lags behind our proposed PerTA in terms of unlearning capability (as measured by FQ and the ES metric on the forget set). The \textit{Pruning} method performs well on simple tasks, such as the 1\% unlearning setting, but its performance drops sharply as the task difficulty increases with higher unlearning ratios. The \textit{SoftMax} method is able to achieve both forgetting and retention, yet it remains inferior to \textit{PerTA-grad} and \textit{PerTA-fisher}. In addition, the results indicate that estimating gradients on $\theta_{\rm full}$ or $\theta_{\rm 0}$ leads to negligible differences in performance.

\textbf{Detailed Results of the General Form.} Considering the general form of $f_{\rm oprt}(A, B)=|A|^{\circ \tau}/(|A|^{\circ \tau}+|B|^{\circ \tau})$ but not using the absolute gradient or the diagonal Fisher Information approximation, different $\tau$s can be applied. We conduct experiments for $\tau \in \{0,0.25,0.5,1,2,4,8\}$, with the quantitative results shown in Table~\ref{tab:tau} as a supplement to Figure~\ref{fig:tau}. The cases of $\tau=1,2$ correspond to our PerTA-grad and PerTA-fisher, respectively. 

The results in Table~\ref{tab:tau} lead to conclusions that are consistent with those discussed in the main body of our paper. PerTA-grad and PerTA-fisher strike a balance between forgetting and retaining: among the different $\tau$-based variants, they achieve relatively strong FQ and ES ($\mathcal D_{\rm f}$) while keeping MU and ES ($\mathcal D_{\rm r}$) at a reasonable level.

\begin{table}[t]
\caption{Results using different $\tau $ in $f_{\rm oprt}$ on TOFU tasks (unlearning 1\%, 5\% and 10\%, using Llama-3.2 1B Instruct, Mean ± Std). Ours are \textcolor{black}{\colorbox{gray!30}{highlighted}}.}
\centering
\begin{small}
\resizebox{\textwidth}{!}{
\begin{tabular}{c|>{\centering\arraybackslash}p{2.9cm}|cc|cc|cc|cc}
\toprule
\multicolumn{1}{l|}{Forgetting}             & \multicolumn{1}{c}{Methods}                            & \multicolumn{2}{|c}{FQ↑}     & \multicolumn{2}{|c}{MU↑}   & \multicolumn{2}{|c}{ES($\mathcal D_{\rm f}$)↓} & \multicolumn{2}{|c}{ES($\mathcal D_{\rm r}$)↑} \\
\midrule
\multirow{9}{*}{1\%}  & \multicolumn{1}{c|}{Full}                    & \multicolumn{2}{c|}{-2.170}  & \multicolumn{2}{c|}{0.599} & \multicolumn{2}{c|}{0.743}      & \multicolumn{2}{c}{0.737}     \\
                                 & \multicolumn{1}{c|}{GT}                          & \multicolumn{2}{c|}{0.000}   & \multicolumn{2}{c|}{0.599} & \multicolumn{2}{c|}{0.069}      & \multicolumn{2}{c}{0.751}     \\
                                 
                                 & $\tau=0$    & -1.451 &
±0.131     & 0.587 & ±0.000
      & 0.116 & ±0.000         & 0.611 & ±0.004        \\
                                 & $\tau=0.25$ & -0.089         & ±0.037     & 0.577       & ±0.000      & 0.095         & ±0.001         & 0.505         & ±0.001        \\
                                 & $\tau=0.5$  & -0.197         & ±0.057     & 0.579       & ±0.001      & 0.095         & ±0.001         & 0.522         & ±0.002        \\
                                 \rowcolor{gray!30}& $\tau=1 (\text{grad})$  & -0.289         & ±0.073     & 0.581       & ±0.001      & 0.075         & ±0.000         & 0.551         & ±0.001        \\
                                 \rowcolor{gray!30}& $\tau=2 (\text{fisher})$  & -0.576         & ±0.000     & 0.586       & ±0.000      & 0.085         & ±0.001         & 0.600         & ±0.001        \\
                                 & $\tau=4$  & -1.013         & ±0.000     & 0.591       & ±0.000      & 0.094         & ±0.001         & 0.650         & ±0.003        \\
                                 & $\tau=8$  & -1.544         & ±0.000     & 0.598       & ±0.001      & 0.276         & ±0.004         & 0.700         & ±0.001        \\

\midrule
\multirow{9}{*}{5\%}  & \multicolumn{1}{c|}{Full}                    & \multicolumn{2}{c|}{-11.845} & \multicolumn{2}{c|}{0.599} & \multicolumn{2}{c|}{0.727}      & \multicolumn{2}{c}{0.737}     \\
                                 & \multicolumn{1}{c|}{GT}                          & \multicolumn{2}{c|}{0.000}   & \multicolumn{2}{c|}{0.599} & \multicolumn{2}{c|}{0.063}      & \multicolumn{2}{c}{0.746}     \\
                                
                                 & $\tau=0$    & -1.253 & ±0.237     & 0.560 & ±0.001      & 0.090 & ±0.004         & 0.396  & ±0.004        \\
                                 & $\tau=0.25$ & -0.784         & ±0.090     & 0.543       & ±0.002      & 0.068         & ±0.001         & 0.299         & ±0.008        \\
                                 & $\tau=0.5$  & -0.754         & ±0.132     & 0.545       & ±0.001      & 0.069         & ±0.002         & 0.300         & ±0.008        \\
                                 \rowcolor{gray!30}& $\tau=1 (\text{grad})$  & -0.661         & ±0.125     & 0.546       & ±0.001      & 0.069         & ±0.002         & 0.310         & ±0.007        \\
                                 \rowcolor{gray!30}& $\tau=2 (\text{fisher})$  & -0.339         & ±0.115     & 0.553       & ±0.001      & 0.077         & ±0.002         & 0.348         & ±0.002        \\
                                 & $\tau=4$  & -0.933         & ±0.292     & 0.565       & ±0.001      & 0.092         & ±0.001         & 0.420         & ±0.004        \\
                                 & $\tau=8$  & -7.008         & ±0.319     & 0.581       & ±0.001      & 0.141         & ±0.008         & 0.573         & ±0.004        \\

                                 \midrule
\multirow{9}{*}{10\%} & \multicolumn{1}{c|}{Full}                    & \multicolumn{2}{c|}{-21.408} & \multicolumn{2}{c|}{0.599} & \multicolumn{2}{c|}{0.706}      & \multicolumn{2}{c}{0.737}     \\
                                 & \multicolumn{1}{c|}{GT}                          & \multicolumn{2}{c|}{0.000}   & \multicolumn{2}{c|}{0.591} & \multicolumn{2}{c|}{0.059}      & \multicolumn{2}{c}{0.746}     \\
                                
                                 & $\tau=0$    & -2.757 & ±0.189     & 0.548 & ±0.002      & 0.082 & ±0.001         & 0.309 & ±0.007        \\
                                 & $\tau=0.25$ & -1.270         & ±0.135     & 0.539       & ±0.002      & 0.071         & ±0.001         & 0.262         & ±0.004        \\
                                 & $\tau=0.5$  & -1.186         & ±0.066     & 0.541       & ±0.002      & 0.072         & ±0.002         & 0.265         & ±0.004        \\
                                 \rowcolor{gray!30}& $\tau=1 (\text{grad})$  & -1.107         & ±0.064     & 0.542       & ±0.001      & 0.071         & ±0.002         & 0.266         & ±0.003        \\
                                 \rowcolor{gray!30}& $\tau=2 (\text{fisher})$  & -1.686         & ±0.265     & 0.548       & ±0.001      & 0.077         & ±0.002         & 0.295         & ±0.006        \\
                                 & $\tau=4$  & -3.490         & ±0.210     & 0.558       & ±0.001      & 0.088         & ±0.002         & 0.354         & ±0.009        \\
                                 & $\tau=8$  & -9.796         & ±0.462     & 0.575       & ±0.001      & 0.138         & ±0.000         & 0.493         & ±0.003  \\     

                                \bottomrule
\end{tabular}}

\end{small}
\label{tab:tau}
\end{table}

\textbf{Detailed Results of Running Time.} As a supplement to Figure~\ref{fig:time}, Table~\ref{tab:time} shows the quantitative runtime comparison between the best-performing training-based method, GD and NPO+, and our PerTA. In contrast to training-based methods that demand multiple iterations, the runtime of PerTA can be broken down into three components: obtaining $\theta_{\rm fgt}$, computing $W$, and performing task arithmetic, with the last step being negligible (0.0002 min in Table~\ref{tab:time}). PerTA thus inherits the efficiency of task arithmetic, yielding substantial runtime savings-a benefit that becomes increasingly evident as task complexity rises (i.e., when unlearning larger proportions). Furthermore, as demonstrated earlier, competitive performance can already be achieved by estimating gradients with only 20\% of the data, indicating additional potential for reducing runtime. Collectively, these observations underscore the high time efficiency of PerTA.
\begin{table}
\caption{Time comparison of the best-performing training-based method GD, NPO+ and our PerTA ((min), unlearning 1\%, 5\% and 10\%, Llama-3.2 1B Instruct).}
\vspace{0.1cm}
\centering
\begin{small}
\resizebox{\textwidth}{!}{
\begin{tabular}{>{\centering\arraybackslash}p{1.6cm}|c|ccc|c}
\toprule
                      Forgetting &        Methods              & Getting $\theta_{\rm fgt}$       & Calculating $W_{\rm grad|fisher}$ & Task Arithmetic           & Total \\
                      \midrule
\multirow{6}{*}{1\%}  & GD                   & -                       & -                                 & -                       & 3.4673  \\
                      & NPO+                 & -                       & -                                 & -                       & 4.6557  \\
                      & PerTA (grad)           & \multirow{4}{*}{0.3944} & 2.0207                            & \multirow{4}{*}{0.0002} & 2.4153  \\
                      & PerTA (fisher)         &                         & 2.2118                            &                         & 2.6064  \\
                      & PerTA (grad) w/ 20\%   &                         & 0.4528                            &                         & 0.8474  \\
                      & PerTA (fisher) w/ 20\% &                         & 0.4188                            &                         & 0.8134  \\
                      \midrule
\multirow{6}{*}{5\%}  & GD                   & -                       & -                                 & -                       & 5.2072  \\
                      & NPO+                 & -                       & -                                 & -                       & 12.3739 \\
                      & PerTA (grad)           & \multirow{4}{*}{2.3918} & 2.0253                            & \multirow{4}{*}{0.0002} & 4.4172  \\
                      & PerTA (fisher)         &                         & 2.2201                            &                         & 4.6121  \\
                      & PerTA (grad) w/ 20\%   &                         & 0.4378                            &                         & 2.8297  \\
                      & PerTA (fisher) w/ 20\% &                         & 0.4179                            &                         & 2.8098  \\
                      \midrule
\multirow{6}{*}{10\%} & GD                   & -                       & -                                 & -                       & 7.2508  \\
                      & NPO+                 & -                       & -                                 & -                       & 23.0168 \\
                      & PerTA (grad)           & \multirow{4}{*}{4.8281} & 2.0231                            & \multirow{4}{*}{0.0002} & 6.8514  \\
                      & PerTA (fisher)         &                         & 2.2177                            &                         & 7.0459  \\
                      & PerTA (grad) w/ 20\%   &                         & 0.4368                            &                         & 5.2651  \\
                      & PerTA (fisher) w/ 20\% &                         & 0.4134                            &                         & 5.2416 \\
                      \bottomrule
\end{tabular}}

\end{small}
\label{tab:time}
\end{table}

\subsection{Different Models for Per-parameter Weights}
\label{app:model}
\begin{figure}[t]
    \centering
    \includegraphics[width=0.9\linewidth]{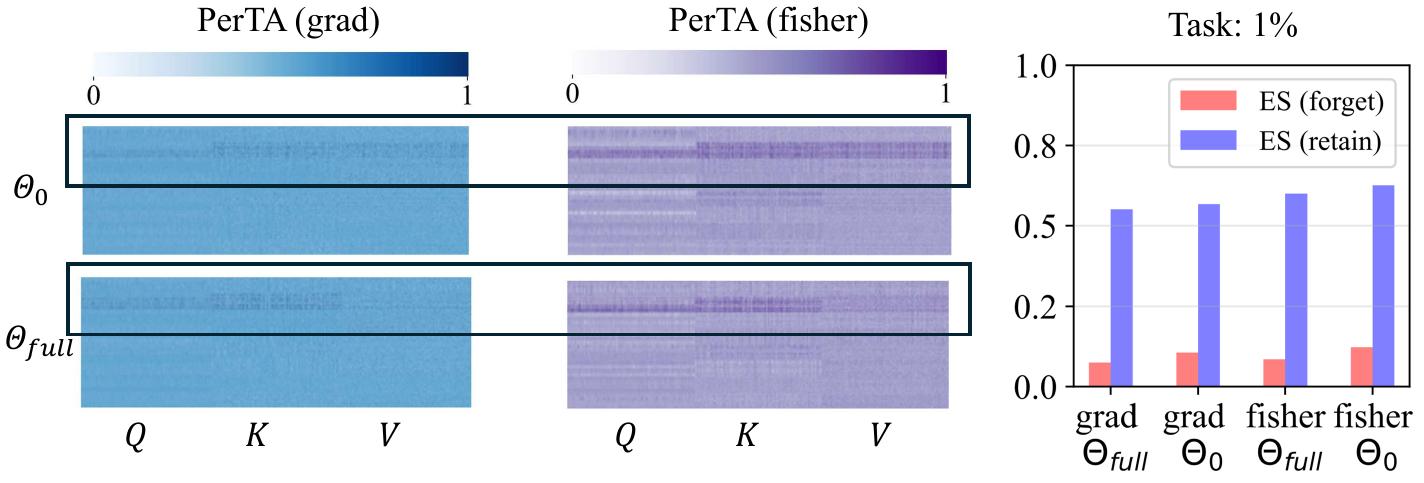}
    \caption{Visualization of $W_{\rm grad}, W_{\rm fisher}$ for parameters in the last two $Q, K, V$ attention layers (left), and corresponding ES on forget and retain sets (right), when employing $\theta_0$ or $\theta_{\rm full}$ to estimate $W_{\rm grad}, W_{\rm fisher}$ (unlearning 1\% on TOFU, using Llama-3.2 1B Instruct).}
    \label{fig:visualization}
\end{figure}
To further illustrate the difference between using $\theta_0$ (the retained LLM) and $\theta_{\rm full}$ (the finetuned LLM) to predict $W$ shown in Table~\ref{tab:ablation}, Figure~\ref{fig:visualization} presents a comparison. The left side of Figure~\ref{fig:visualization} visualizes the weight magnitudes of $W$ (predicted by $\theta_0$ and $\theta_{\rm full}$, respectively) corresponding to the $Q$, $K$, and $V$ matrices in the last two attention layers, while the right side reports the corresponding ES scores in bar plots. From the visualizations on the left, we observe that both PerTA-grad and PerTA-fisher exhibit highly similar patterns regardless of whether $W$ is predicted by $\theta_0$ and $\theta_{\rm full}$ (highlighted by the black boxes). This indicates that the key parameters--those with large weights--are largely consistent across the two predictors, and vice versa. On the right, the ES results confirm this observation: the numerical metrics are very close, consistent with Table~\ref{tab:ablation}. 

These findings suggest that either $\theta_0$ or $\theta_{\rm full}$ can be used to predict $W$, with negligible differences. A plausible explanation is that the gap between the pretrained model and the finetuned model is relatively small. This conclusion further supports the applicability of PerTA to post-training models, thereby broadening its range of use cases.
\subsection{Visualization Results of Weights}
\label{app:visual}
\begin{figure}[!h]
    \centering
    \includegraphics[width=\linewidth]{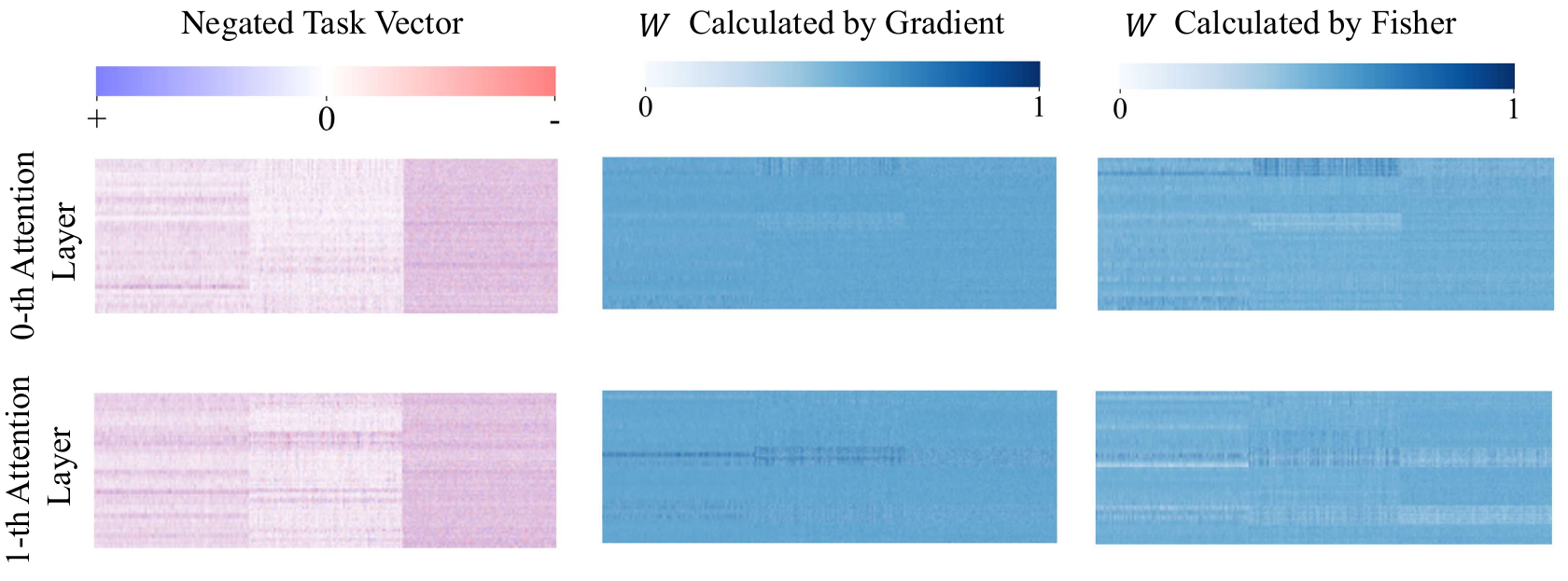}
    \caption{Visualization of TV, $W_{\rm grad}, W_{\rm fisher}$ for parameters in the 0-th, 1-st $Q, K, V$ attention layers (unlearning 1\% on TOFU, using Llama-3.2 1B Instruct).}
    \label{fig:vis_first}
\end{figure}
\begin{figure}[!h]
    \centering
    \includegraphics[width=\linewidth]{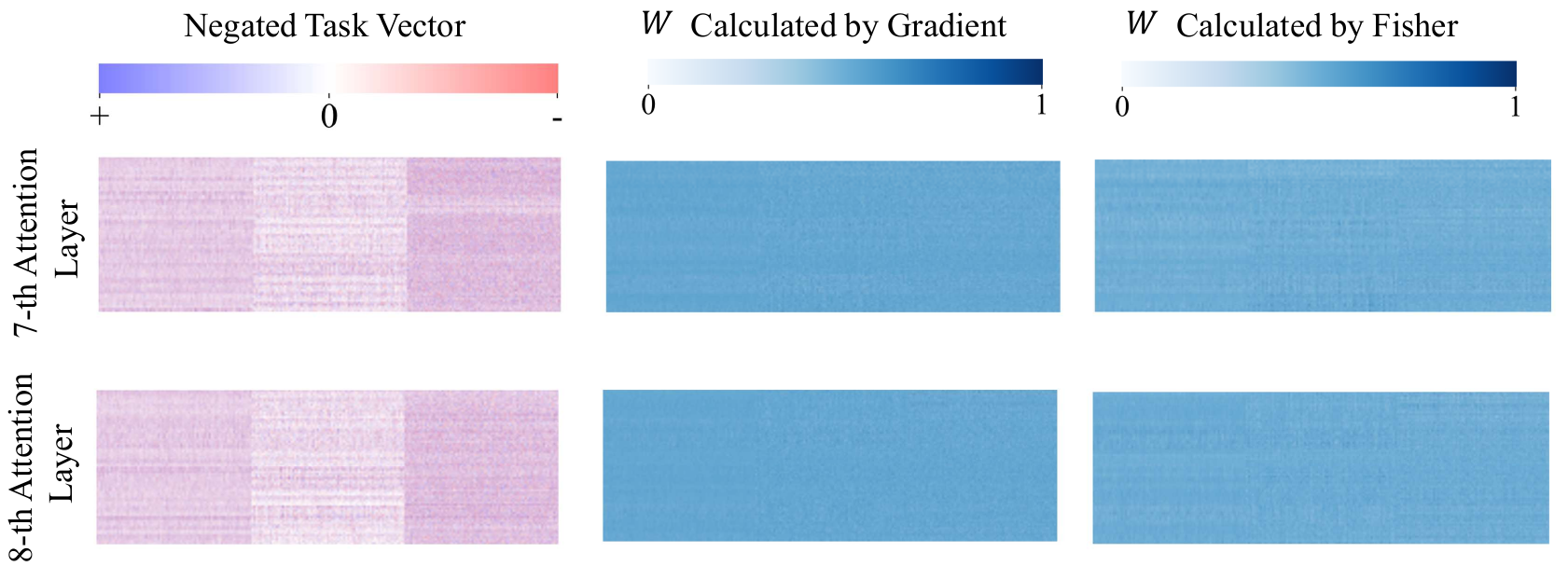}
    \caption{Visualization of TV, $W_{\rm grad}, W_{\rm fisher}$ for parameters in the 7-th, 8-th $Q, K, V$ attention layers (unlearning 1\% on TOFU, using Llama-3.2 1B Instruct).}
    \label{fig:vis_mid}
\end{figure}
\begin{figure}[!h]
    \centering
    \includegraphics[width=\linewidth]{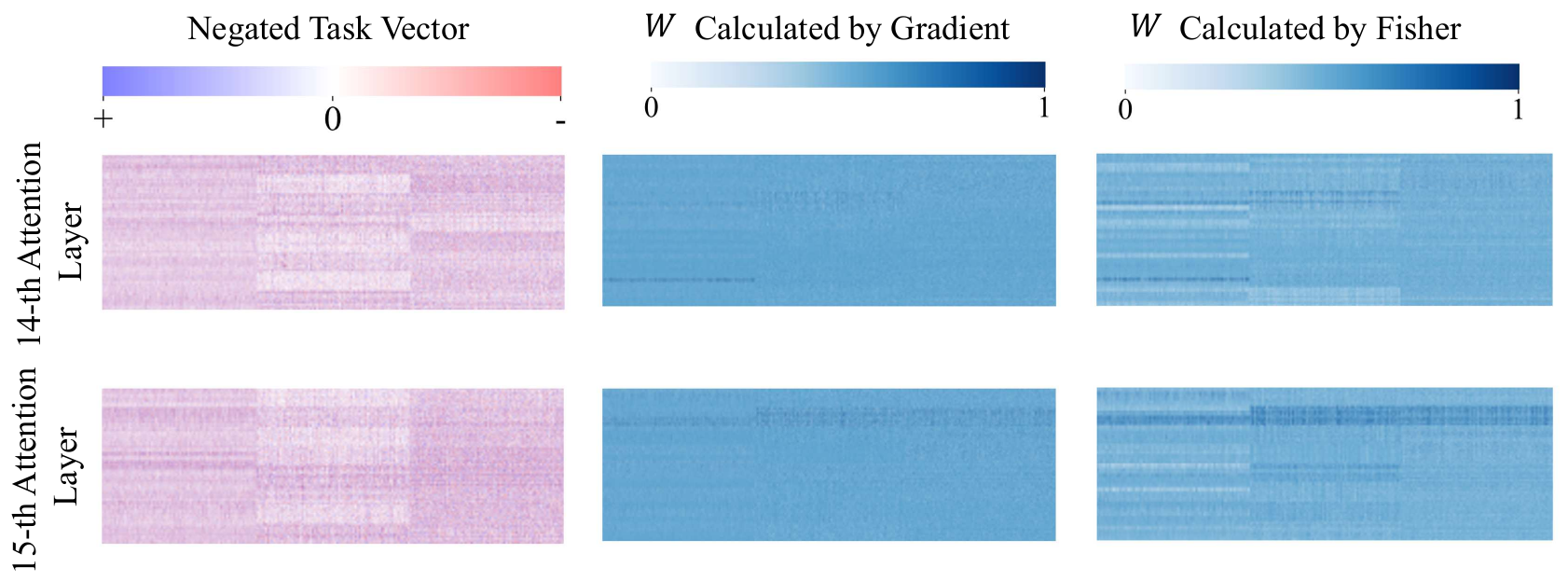}
    \caption{Visualization of TV, $W_{\rm grad}, W_{\rm fisher}$ for parameters in the 14-th, 15-th $Q, K, V$ attention layers (unlearning 1\% on TOFU, using Llama-3.2 1B Instruct).}
    \label{fig:vis_last}
\end{figure}
Figures~\ref{fig:vis_first}-\ref{fig:vis_last} visualize the weight magnitudes of the $Q$, $K$, and $V$ matrices in the shallow, middle, and final attention layers of the LLM for both TV and $W$. For TV, we observe that the weight magnitudes increase progressively from shallow to deeper layers, indicating that the magnitude of parameter changes induced by unlearning grows with layer depth. 

In contrast, the analysis of $W$ may provide insight into the layer-wise sensitivity of LLM parameters to the differences between forget and retain data. We highlight two key observations. First, compared to PerTA-grad, PerTA-fisher exhibits more pronounced weight differences (as evidenced by the larger contrast between light and dark regions in Figure~\ref{fig:vis_first}-\ref{fig:vis_last}). This is because PerTA-fisher relies on the squared gradients rather than the raw gradients, thereby amplifying the differences between the forget and retain sets. In practice, however, both PerTA-grad and PerTA-fisher yield similar performance on the evaluation metrics, suggesting that either variant can be employed effectively.

Second, relative to the middle layers of the LLM, the initial and final layers contain more weights close to the extremes (i.e., near 0 or 1). This implies that parameters in the shallow and final layers are more sensitive to gradient differences between the forget and retain sets. Interestingly, this aligns with prior findings on LLM representations~\citep{layers1,layers2}: shallow layers primarily capture surface features (e.g., words, subwords, positional information), middle layers encode syntactic features, and final layers specialize in semantic features. The results in Figures~\ref{fig:vis_first}-\ref{fig:vis_last} are consistent with this interpretation. Specifically, surface and semantic features exhibit greater discrepancies between forget and retain sets (e.g., TOFU involves differences in author names, domain-specific terminology, and deeper semantic associations with personal information), whereas syntactic structures remain largely unaffected. Consequently, our flexible PerTA assigns larger weight differences to parameters in the shallow and final layers. This insight suggests a potential future direction for further optimization: pruning or fixing selected middle layers to reduce computational overhead without sacrificing performance.
\begin{table}[!ht]
\centering
\caption{{Results of different methods on unlearning 5\% of TOFU, using Llama-3.2 8B as the pretrained model. }}
\centering
\begin{small}

\begin{tabular}{c|cccccc}
\toprule
                 & \multicolumn{1}{c}{FQ↑}                 & \multicolumn{1}{c}{MU↑}                & \multicolumn{1}{c}{ES($D_f$)↓} & \multicolumn{1}{c}{ES($D_r$)↑} & \multicolumn{1}{c}{ES($D_r$)-ES($D_f$)↑} & \multicolumn{1}{c}{Gib↑} \\
\midrule
Full (reference) & \multicolumn{1}{c}{-12.184}             & \multicolumn{1}{c}{0.628}              & \multicolumn{1}{c}{0.972}     & \multicolumn{1}{c}{0.992}     & 0.020                                  & 0.852                    \\
GT (reference)   & \multicolumn{1}{c}{0.000}               & \multicolumn{1}{c}{0.632}              & \multicolumn{1}{c}{0.074}     & \multicolumn{1}{c}{0.992}     & 0.918                                  & 0.886                    \\
\midrule
GA               & -118.712                                & 0.000                                  & 0.033                         & 0.035                         & 0.002                                  & 0.038                    \\
GD               & -10.225                                 & 0.509                                  & 0.158                         & 0.397                         & 0.239                                  & 0.811                    \\
NPO              & -11.183                                 & 0.131                                  & 0.033                         & 0.037                         & 0.004                                  & 0.141                    \\
NPO+             & -7.888                                  & 0.569                                  & 0.160                         & 0.521                         & 0.361                                  & \textbf{0.914}           \\
PerTA-grad (ours)  & \cellcolor[HTML]{FFFFFF}\textbf{-4.529} & \cellcolor[HTML]{FFFFFF}\textbf{0.659} & \cellcolor[HTML]{FFFFFF}0.164 & \cellcolor[HTML]{FFFFFF}0.882 & \textbf{0.718}                         & 0.895   \\
\bottomrule
\end{tabular}\label{tab:res_8B}

\end{small}
\end{table}

\begin{table}[!ht]
\centering
\caption{Results of different methods on unlearning 5\% of TOFU, using Phi-3.5 as the pretrained model.}
\centering
\begin{small}

\begin{tabular}{c|cccccc}
\toprule
                 & FQ↑             & MU↑            & ES($D_f$)↓ & ES($D_r$)↑ & ES($D_r$)-ES($D_f$)↑ & Gib↑           \\
\midrule
Full (reference) & -13.232         & 0.693          & 0.868     & 0.835     & -0.033             & 0.866          \\
GT (reference)   & 0.000           & 0.678          & 0.082     & 0.855     & 0.773              & 0.881          \\
\midrule
GA               & -11.511         & 0.073          & 0.027     & 0.028     & 0.001              & 0.822          \\
GD               & -11.183         & 0.665          & 0.344     & 0.574     & 0.231              & 0.875          \\
NPO              & -12.877         & 0.278          & 0.538     & 0.594     & 0.057              & 0.855          \\
NPO+             & -10.859         & 0.552          & 0.591     & 0.761     & 0.170              & 0.877          \\
PerTA-grad (ours)  & \textbf{-3.548} & \textbf{0.667} & 0.107     & 0.412     & \textbf{0.305}     & \textbf{0.879} \\
\bottomrule
\end{tabular}\label{tab:res_phi}

\end{small}
\end{table}

\begin{table}[!h]
\centering
\caption{Average results of PerTA with quantization attacks on TOFU 1\%, 5\%, 10\% unlearning tasks.}
\begin{small}

\begin{tabular}{c|cccccc}
\toprule
                           & FQ↑             & MU↑            & ES($D_f$)↓ & ES($D_r$)↑ & ES($D_r$)-ES($D_f$)↑ & Gib↑           \\
\midrule
Full                       & -11.808         & 0.599          & 0.726      & 0.737      & 0.011                & 0.871          \\
GT                         & 0.000           & 0.596          & 0.064      & 0.748      & 0.684                & 0.894          \\
\midrule
GA                         & -81.114         & 0.199          & 0.086      & 0.244      & 0.157                & 0.484          \\
GD                         & -8.720          & 0.491          & 0.112      & 0.295      & 0.183                & 0.789          \\
NPO                        & -4.842          & 0.198          & 0.086      & 0.246      & 0.160                & 0.592          \\
NPO+                       & -3.528          & 0.493          & 0.122      & 0.316      & 0.194                & 0.911          \\
PerTA-grad (ours) w/o attack & \textbf{-0.686} & 0.556          & 0.072      & 0.376      & 0.304                & \textbf{0.915} \\
PerTA-grad (ours) w/ attack  & -1.340          & \textbf{0.560} & 0.095      & 0.421      & \textbf{0.325}       & 0.909         \\
\bottomrule
\end{tabular}\label{tab:quant_attack}

\end{small}
\end{table}

\subsection{Results on Larger Models and Alternative LLM Families}
\label{app:other}
Tables~\ref{tab:res_8B} and Table~\ref{tab:res_phi} present our method’s performance on larger models and on models from other LLM families. The results indicate that our PerTA exhibits good generalization ability: it achieves competitive unlearning performance even when applied to larger models and different types of LLMs.
\subsection{Results of Quantization Attacks}
\label{app:attack}
Some recent research~\citep{zhangcatastrophic} have found that applying quantization to models that have undergone unlearning can restore the "forgotten" information. Therefore, conducting attack experiments on PerTA to reveal whether it possesses robustness is crucial. 

Accordingly, we evaluate the model after unlearning--using Llama-3.2 1B as an example--and the results are shown in Table~\ref{tab:quant_attack}. The results show that, fortunately, the impact of quantization on PerTA is limited, and PerTA still outperforms other methods after the attack.

\end{document}